\documentclass[acmtog,nonacm,balance=false]{acmart}
\AtBeginDocument{%
  \providecommand\BibTeX{{%
    \normalfont B\kern-0.5em{\scshape i\kern-0.25em b}\kern-0.8em\TeX}}}

\newcommand{\revAdded}[1]{{#1}}
\newcommand{\revRemoved}[1]{{}}
\newcommand{\revReplaced}[2]{#2}

\newcommand{\GenericNetworkParameters}{\theta}
\newcommand{\GenericNetwork}{\textrm{NN}_\GenericNetworkParameters}
\newcommand{\GeometryNetwork}{\textrm{NN}^{\textrm{geo}}_\GenericNetworkParameters}

\newcommand{\RadianceNetwork}{\textrm{NN}^{\textrm{rad}}_\GenericNetworkParameters}
\newcommand{\SDF}{f}
\newcommand{\KernelSize}{s}
\newcommand{\LengthAlongRay}{\tau}
\newcommand{\Color}{\mathbf{c}}
\newcommand{\Density}{\sigma}

\newcommand{\Position}{\mathbf{x}}
\newcommand{\Direction}{\mathbf{d}}
\newcommand{\Ray}{\mathbf{r}}
\newcommand{\RayOrigin}{\mathbf{o}}
\newcommand{\RayDirection}{\mathbf{d}}
\newcommand{\SetOfRays}{\mathcal{R}}
\newcommand{\NumberOfSamples}{N}
\newcommand{\SetOfSamples}{\mathcal{X}}
\newcommand{\StepSize}{\delta}
\newcommand{\NoiseSTD}{\varepsilon}

\newcommand{\ErodedField}{\textrm{SDF}_-}
\newcommand{\DilatedField}{\textrm{SDF}_+}
\newcommand{\LevelsetVelocity}{v}

\newcommand{\DensityMin}{\Density_\textrm{min}}
\newcommand{\DensityDilateVelCoef}{\beta_d}
\newcommand{\ErodeVelocityMax}{v_\textrm{max}}
\newcommand{\DensityErodeVelCoef}{\beta_e}
\newcommand{\InnerMesh}{\textsf{M}_-}
\newcommand{\OuterMesh}{\textsf{M}_+}

\newcommand{\PosEncoding}{\Psi}
\newcommand{\DirEncoding}{\gamma}
\newcommand{\Loss}{\mathcal{L}}
\newcommand{\LossWeight}{\lambda}

\newcommand{\GeoFeature}{\mathbf{f}_\textrm{geo}}
\newcommand{\NormalVector}{\mathbf{n}}

\definecolor{gold}{RGB}{221, 196, 65}
\definecolor{silver}{RGB}{215, 215, 215}
\definecolor{bronze}{RGB}{126, 66, 5}
\definecolor{green_zoomin}{RGB}{42, 137, 0}
\definecolor{nsteps_blue}{rgb}{0.03137254901960784, 0.18823529411764706, 0.4196078431372549}

\newcommand{\tikzcircle}[2][red,fill=red]{\tikz[baseline=-0.7ex]\draw[#1,radius=#2] (0,0) circle ;}%

\newcommand{\IntervalWidth}{w}
\newcommand{\IntervalSampleSpacing}{\delta_s}

\newcommand{\NIntervalSamplesMax}{N_\textrm{max}}

\newcommand{\IntervalSingleSampleWidth}{w_\textrm{s}}

\newcommand{\eg}{\emph{e.g.}} %
\newcommand{\vs}{\emph{vs.}} %
\newcommand{\etal}{\emph{et al.}} %

\renewcommand{\eqref}[1]{Equation~\ref{eq:#1}}
\newcommand{\figref}[1]{Figure~\ref{fig:#1}}
\newcommand{\secref}[1]{Section~\ref{sec:#1}}

\newcommand{\tabref}[1]{Table~\ref{tab:#1}}

\newcommand{\DTUDataset}{\emph{DTU}} %
\newcommand{\ShellyDataset}{\emph{Shelly}} %
\newcommand{\NeRFDataset}{\emph{NeRFSynthetic}} %
\newcommand{\MipNerfDataset}{\emph{MipNeRF360}} %

\newcommand{\PSNR}{\emph{PSNR}} %
\newcommand{\LPIPS}{\emph{LPIPS}} %
\newcommand{\SSIM}{\emph{SSIM}} %

\acmSubmissionID{774}

\usepackage{multirow}
\usepackage{enumitem}
\usepackage{amsmath}
\usepackage{amsfonts}
\usepackage{nicefrac}
\usepackage{mathtools}
\usepackage{wrapfig}
\usepackage{tikz}
\usepackage{xcolor}
\usepackage{colortbl}
\usepackage{adjustbox}
\usepackage{booktabs} %
\usepackage{tabularx}
\usepackage{xspace}
\usepackage{ulem}
\newcolumntype{Y}{>{\centering\arraybackslash}X}
\newcolumntype{C}[1]{>{\centering\arraybackslash}p{#1}}

\usepackage{custom_style} %


\begin{document}

\title{\vspace*{4mm}Adaptive Shells for Efficient Neural Radiance Field Rendering\vspace*{1mm}}

\author{Zian Wang}
\orcid{0000-0003-4166-3807}
\authornote{Authors contributed equally.}
\affiliation{%
  \institution{NVIDIA, University of Toronto, Vector Institute}
  \city{Toronto}
  \country{Canada}
}
\email{zianw@nvidia.com}

\author{Tianchang Shen}
\authornotemark[1]
\orcid{0000-0002-7133-2761}
\affiliation{%
  \institution{NVIDIA, University of Toronto, Vector Institute}
   \city{Toronto}
   \country{Canada}
  }
\email{frshen@nvidia.com}

\author{Merlin Nimier-David}
\orcid{0000-0002-6234-3143}
\authornotemark[1]
\affiliation{%
  \institution{NVIDIA}
   \city{Z\"urich}
   \country{Switzerland}
  }
\email{mnimierdavid@nvidia.com}

\author{Nicholas Sharp}
\orcid{0000-0002-2130-3735}
\affiliation{%
  \institution{NVIDIA}
   \city{Seattle}
   \country{USA}
  }
\email{nsharp@nvidia.com}

\author{Jun Gao}
\orcid{0000-0002-3521-0417}
\affiliation{%
  \institution{NVIDIA, University of Toronto, Vector Institute}
   \city{Toronto}
   \country{Canada}
  }
\email{jung@nvidia.com}

\author{Alexander Keller}
\orcid{0000-0002-9144-5982}
\affiliation{%
  \institution{NVIDIA}
   \city{Berlin}
   \country{Germany}
  }
\email{akeller@nvidia.com}

\author{Sanja Fidler}
\orcid{0000-0003-1040-3260}
\affiliation{%
  \institution{NVIDIA, University of Toronto, Vector Institute}
   \city{Toronto}
   \country{Canada}
  }
\email{sfidler@nvidia.com}

\author{Thomas M\"uller}
\orcid{0000-0001-7577-755X}
\affiliation{%
  \institution{NVIDIA}
   \city{Z\"urich}
   \country{Switzerland}
  }
\email{tmueller@nvidia.com}

\author{Zan Gojcic}
\orcid{0000-0001-6392-2158}
\affiliation{%
  \institution{NVIDIA}
   \city{Z\"urich}
   \country{Switzerland}
  }
\email{zgojcic@nvidia.com}

\renewcommand{\shortauthors}{Wang, Shen, Nimier-David, \etal{}}

\acmSubmissionID{774}



\begin{abstract}
Neural radiance fields achieve unprecedented quality for novel view synthesis, but their volumetric formulation remains expensive, requiring a huge number of samples to render high-resolution images.
Volumetric encodings are essential to represent fuzzy geometry such as foliage and hair, and they are well-suited for stochastic optimization.
Yet, many scenes ultimately consist largely of solid surfaces which can be accurately rendered by a single sample per pixel.
Based on this insight, we propose a neural radiance formulation that smoothly transitions between volumetric- and surface-based rendering, greatly accelerating rendering speed and even improving visual fidelity.
Our method constructs an explicit mesh envelope which spatially bounds a neural volumetric representation.
In solid regions, the envelope nearly converges to a surface and can often be rendered with a single sample.
To this end, we generalize the NeuS~\cite{wang2021neus} formulation with a learned spatially-varying kernel size which encodes the spread of the density, fitting a wide kernel to volume-like regions and a tight kernel to surface-like regions.
We then extract an explicit mesh of a narrow band around the surface, with width determined by the kernel size, and fine-tune the radiance field within this band.
At inference time, we cast rays against the mesh and evaluate the radiance field only within the enclosed region, greatly reducing the number of samples required.
Experiments show that our approach enables efficient rendering at very high fidelity.
We also demonstrate that the extracted envelope enables downstream applications such as animation and simulation.
\vspace{3mm}
\end{abstract}

\begin{teaserfigure}
\centering
\vspace*{-4em}
\includegraphics{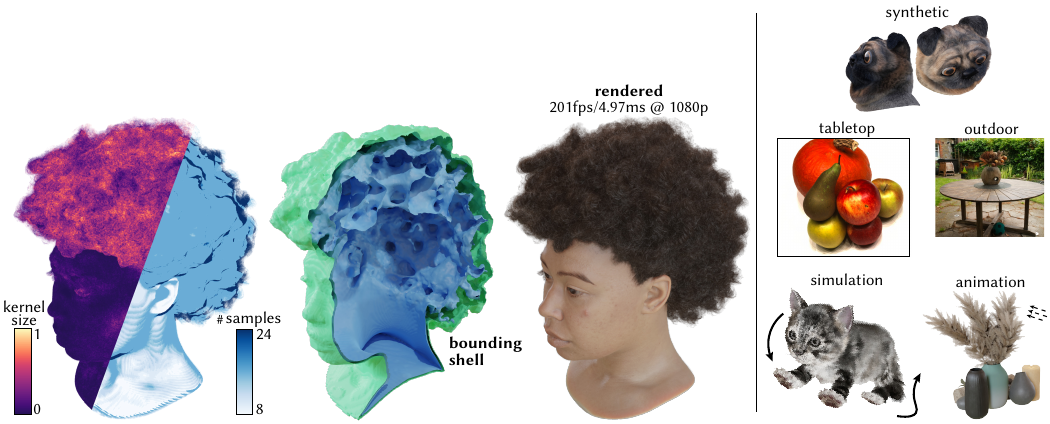}
\caption{
This work presents an approach for efficiently rendering neural radiance fields by restricting volumetric rendering to a narrow band around the object.
\emph{Left:} We first fit a dense neural volume using a new spatially-varying kernel that automatically adapts to be large in volumetric regions such as hair or grass, and small in sharp-surface regions such as skin or furniture. We then extract an explicit bounding mesh of the region to be rendered whose width is determined by the kernel, and render at \revReplaced{interactive}{real-time} rates.
\emph{Right:} the proposed method is general and effective across a wide range of data and well-suited for downstream applications such as simulation and animation.
The face model of the Khady synthetic human shown \emph{left} is courtesy of texturing.xyz.
\label{fig:teaser}
}
\vspace*{1em}
\end{teaserfigure}

\maketitle

\section{Introduction}
\label{sec:Introduction}

Neural radiance fields, which we will refer to as \emph{NeRFs}, have recently emerged as a powerful 3D representation enabling photorealistic novel-view synthesis and reconstruction.
Unlike traditional explicit methods for novel-view synthesis, NeRFs forego high-quality mesh reconstruction and explicit surface geometry in favor of neural networks, which encode the volumetric density and appearance of a scene as a function of 3D spatial coordinates and viewing direction.
However, the high visual fidelity of NeRFs comes at a great computational cost, as the volume rendering formulation requires a large number of samples along each ray and ultimately prevents real-time synthesis of high-resolution novel views.
In tandem, explicit reconstruction and novel-view synthesis have continued to make great progress by leveraging advances in inverse rendering and data-driven priors, but a fidelity gap remains.
The goal of this work is to close this gap by developing a neural volumetric formulation that leverages explicit geometry to accelerate performance, without sacrificing quality.

Much recent and concurrent work has likewise sought to improve the efficiency of NeRF representations and volume rendering.
An important step towards this goal was the evolution from a global large multi-layer perceptron (MLP) representation~\cite{mildenhall2020nerf} to local sparse feature fields combined with shallow MLP decoders~\cite{yu_and_fridovichkeil2021plenoxels, mueller2022ingp}.
This resulted in several orders-of-magnitude speed-ups.
Complementary research to improve the efficiency of NeRFs proposed replacing the neural networks by simpler functions such as spherical harmonics, or baking the volumetric representation onto proxy geometry that accelerates rendering~\cite{yariv2023bakedsdf,chen2022mobilenerf}.
The latter formulation enables especially large speedups and facilitates real-time rendering even on commodity devices~\cite{chen2022mobilenerf}. Yet, doing so compromises the quality as the scene content is projected onto proxy geometry.

In this work, we instead aim to make NeRF rendering more efficient while maintaining or even improving the perceptual quality.
To this end, we propose a narrow-band rendering formulation that enables efficient novel-view synthesis, while enjoying the desirable properties of the volumetric representation (\figref{teaser} \emph{left}).
Our method is inspired by the insight that different regions of the scene benefit from different styles of rendering.
Indeed, fuzzy surfaces with intricate geometry and complex transparency patterns benefit greatly from exhaustive volume rendering, while conversely, opaque smooth surfaces can be well---or potentially even better---represented by a single sample where the ray intersects the surface.
This observation allows us to better distribute the computational cost across the rays by assigning as many samples as needed to faithfully represent the ground-truth appearance.

\begin{figure}[t]
    \centering
    \vspace*{1em}
    \includegraphics{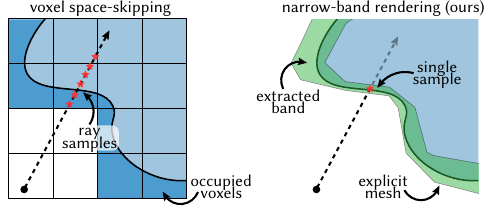}
    \caption{
        One state-of-the-art approach to accelerate volumetric rendering is to skip empty voxels, however this still requires multiple samples within occupied voxels (\emph{left}).
        Our approach extracts a narrow band mesh, for which a single sample at the midpoint is a very good approximation of the surface (\emph{right}).
        \label{fig:surface_vs_voxel}
    }
    \vspace*{1em}
\end{figure}

With the introduction of auxiliary acceleration data structures that promote empty space skipping~\cite{mueller2022ingp}, NeRFs can already render images with a varying number of samples per ray. Still, there remain many challenges that prevent the current formulations from efficiently adapting to the local complexity of the scene (\figref{surface_vs_voxel}).
First, the memory footprint of grid-based acceleration structures scales poorly with resolution.
Second, the smooth inductive bias of MLPs hinders learning a sharp impulse or step function for volume density, and even if such an impulse was learned it would be difficult to sample it efficiently.
Finally, due to the lack of constraints, the implicit volume density field fails to accurately represent the underlying surfaces~\cite{wang2021neus}, which often limits their application in downstream tasks that rely on mesh extraction.

To remedy the last point, ~\cite{wang2021neus, yariv2021volsdf, wang2022neus2} propose to optimize a signed distance function (SDF) along with a kernel size encoding the spread of the density, rather than optimizing density directly.
While this is effective for improving surface representations, the use of a global kernel size contradicts the observation that different regions of the scene demand adaptive treatment.

To address the above challenges, we propose a new volumetric neural radiance field representation. In particular:
\textbf{i}) We generalize the NeuS~\cite{wang2021neus} formulation with a \emph{spatially-varying} kernel width that converges to a wide kernel for fuzzy surfaces, while collapsing to an impulse function for solid opaque surfaces without additional supervision. This improvement alone results in an increased rendering quality across all scenes in our experiments.
\textbf{ii}) We use the learned spatially-varying kernel width to extract a mesh envelope of a narrow band around the surface. The width of the extracted envelope adapts itself to the complexity of the scene and serves as an efficient auxiliary acceleration data structure.
\textbf{iii}) At inference time, we cast rays against the envelope in order to skip empty space and sample the radiance field only in regions which contribute significantly to the rendering. In surface-like regions, the narrow band enables rendering from a single sample, while progressing to a wider kernel and local volumetric rendering for fuzzy surfaces.

The experiments of Section~\ref{sec:experiments} validate the effectiveness of our formulation across several data sets. In addition, the applications of Section~\ref{sec:applications} demonstrate the benefits of our representation.

\begin{figure*}
    \centering
    \vspace*{1em}
    \includegraphics{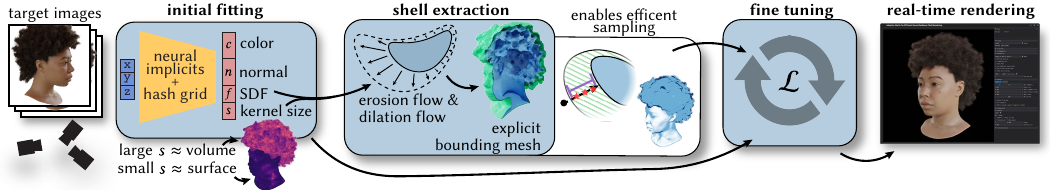}
    \caption{
        Overview of the proposed approach.
        We demonstrate high-fidelity, efficient neural implicit scene reconstruction by efficiently-sampling volumetric rendering inside of an explicit thin shell, which is automatically fit from visual objectives.
        \label{fig:method_overview}
    }
    \vspace*{1em}
\end{figure*}

\section{Related Work}
\label{sec:prev_work}

Synthesizing novel views from a set of images is a longstanding problem in the fields of computer vision and graphics. The classical approaches to novel-view synthesis can be roughly categorized based on the coverage density of the input images. In particular, light field interpolation methods~\cite{Levoy1996lightfields,Gortler1996lumigraph, davis2012ulf} assume that the input views are sampled densely and close to the target view. When the input views are sparse, classical methods usually follow a two-stage approach: In the first stage, they construct a proxy geometry from the images using a combination of a multi-view stereo pipeline~\cite{schoenberger2016sfm, schoenberger2016mvs} and point cloud reconstruction methods~\cite{kazdhan2006psr, kazdhan2013spsr}. In the second stage, the input images are then unprojected onto the geometry either directly in terms of RGB colors~\cite{debevec1996hybrid, wood2000surfacelf, Buehler2001ulr, Fleet2014letcolor} or, more recently, latent features~\cite{Riegler2020FVS, Riegler2021SVS}.
Other lines of research have developed specialized methods for certain classes of objects, such as faces (\eg{} \citet{bi2021deep})---although we show results on synthetic human and animal data (\figref{teaser}), the approach presented here is entirely general.

\paragraph{Neural Radiance Fields (NeRFs)} NeRF~\cite{mildenhall2020nerf} have revolutionized the prevailing paradigm of novel-view synthesis, by using a neural network to represent the scenes as a volumetric (radiance) field that may be queried at any location to return the view-dependent radiance and volume density. \citet{mildenhall2020nerf} synthesize novel views by querying the radiance field along the image rays and accumulating the appearance using volume rendering. The photorealistic quality of NeRF has inspired a large body of follow-up work. NeRF++~\cite{kaizhang2020nerf++} analyzed the difficulties of NeRF to represent unbounded scenes and proposed a background formulation based on the inverted sphere representation. MipNeRF~\cite{barron2021mipnerf} addressed the aliasing effects with an integrated positional encoding. This work was later extended to unbounded scenes~\cite{barron2022mipnerf360} by contracting the volume and using an additional proposal network. ~\cite{kangle2021dsnerf} and \cite{Niemeyer2021Regnerf} tackled the challenging setting with sparse input views and proposed to regularize the volumetric representation using depth supervision or smoothness constraints and data priors based on normalizing flows, respectively. NeRF-W~\cite{martinbrualla2020nerfw} has shown how NeRF can be extended to unstructured collections of images captured in-the-wild, by using per-frame learnable latent codes to compensate for appearance differences and a transient embedding to remove dynamic objects. Alternative representations to neural fields include point clouds~\cite{ruckert2021adop, Kopanas2021pointbasednr}, spheres~\cite{Lassner2021PulsarES}, and 3D Gaussians~\cite{kerbl3Dgaussians}.

\paragraph{Implicit surface representation}
The NeRF formulation has two main shortcomings when it comes to modeling surfaces: i) Besides a lack of regularization of the density field, ii) surface extraction has to be performed at an arbitrary level-set of the density field. In combination, these lead to noisy and low-fidelity surface reconstruction. However, with small changes in the formulation, implicit representations combined with volume rendering~\cite{yariv2020idr, zhang2021PhySG, wang2021neus, yariv2021volsdf, Oechsle2021unisurf, wang2022hfneus} still appear as a promising alternative to classical surface reconstruction approaches from image data~\cite{schoenberger2016mvs}. For example, instead of directly optimizing the density field, ~\cite{wang2021neus, yariv2021volsdf} proposed to decompose it into an SDF and a global kernel size that defines the spread of the density. This allows for extracting accurate surfaces from the zero-level set of the SDF, which can also be regularized using the Eikonal constraint. Similar to NeRFs, implicit surface representations were also combined with local feature fields and auxiliary acceleration data structures~\cite{wang2022neus2, Zhao2022neuralAM, Li2023Neuralangelo, rosu2023permutosdf, tang2023delicate} with the goal of improved efficiency and representation capacity. While our method is built on the NeuS~\cite{wang2021neus} formulation, our main goal is not to improve the accuracy of the extracted surface. Instead, we utilize the SDF to extract a narrow shell that allows us to adapt the representation to the local complexity of the scene and in turn to accelerate rendering.

\paragraph{Accelerating neural volume rendering}
One of the main limitations of NeRFs is the computational complexity of neural volume rendering which slows down both training and inference. Recently, various different directions to accelerate NeRFs have been explored. For example, replacing a global MLP with a (sparse) local feature field combined with a shallow MLP~\cite{liu2020nsvf, yu2021plenoctrees, mueller2022ingp, SunSC22DVGO, Chen2022tensorrf} or the spherical harmonics embedding~\cite{yu_and_fridovichkeil2021plenoxels, karnewar2022relufields, Chen2022tensorrf}, partitioning the volume into a large number of local (shallow) MLPs~\cite{Reiser2021kilonerf, Rebain2021DeRF}, or using efficient sampling strategies~\cite{neff2021donerf, Hu2022EfficientNeRF, lin2022efficientnerf, kurz2022adanerf}, or image-space convolutions~\cite{cao2022real,wan2023Duplex}.
However, even the most optimized volumetric representations~\cite{mueller2022ingp} are still much slower than pure surface-based approaches such as NvDiffRec~\cite{Munkberg2022nvdiffrec}.

To further increase the efficiency of the inference phase, volumetric representations can be baked onto a proxy surface geometry~\cite{chen2022mobilenerf, yariv2023bakedsdf, wan2023Duplex} that can be efficiently rendered using high-performance rasterization pipelines. An alternative \emph{"baking"} strategy is to precompute the outputs of the neural network and store them on a (sparse) discrete grid that acts as a lookup during inference~\cite{hedman2021snerg, reiser2023merf}. In this work, we investigate an alternate approach to speeding up the (volumetric) rendering, by adapting the number of samples required to render each pixel to the underlying local complexity of the scene. Note that our formulation is complementary to the \emph{"baking"} approaches and we consider the combination of both an interesting avenue for future research.

\section{Method}
\label{sec:method}

Our method (see \figref{method_overview}) builds on NeRF~\cite{mildenhall2020nerf} and NeuS~\cite{wang2021neus}.
Specifically, we generalize NeuS~\cite{wang2021neus} with a new spatially-varying kernel (\secref{extending_neus}), which improves the quality and guides the extraction of a narrow-band shell (\secref{shell_extraction}).
Then, the neural representation is fine-tuned (\secref{training}) within the shell that significantly accelerates rendering (\secref{narrow_band_rendering}).

\subsection{Preliminaries}
\label{sec:preliminaries}

NeRF~\cite{mildenhall2020nerf} represents a scene as a volumetric radiance field that maps a 3D point $\Position \in \mathbb{R}^3$ and a viewing direction $\Direction \in \mathbb{R}^3$ to the volume density $\Density$ and the emitted view-dependent color $\Color \in \mathbb{R}^3$.
This volumetric field is represented by a neural network $\GenericNetwork(\cdot)$ with parameters $\GenericNetworkParameters$, such that $(\Color, \Density) = \GenericNetwork(\Position, \Direction)$.
The scene can then be rendered along a ray $\Ray = \RayOrigin + \LengthAlongRay\RayDirection$ with origin $\RayOrigin \in \mathbb{R}^3$ and direction $\RayDirection \in \mathbb{R}^3$ from $\LengthAlongRay_n$ to $\LengthAlongRay_f$ via standard volumetric rendering
\begin{equation}
    \label{eq:vol_rendering_cont}
    \Color(\Ray) = \int_{\LengthAlongRay_n}^{\LengthAlongRay_f} \exp \Big[\int_{\LengthAlongRay_n}^{\LengthAlongRay} -\Density(\Ray(z)) dz  \Big] \Density(\Ray(\LengthAlongRay)) \Color(\Ray(\LengthAlongRay), \RayDirection) d\LengthAlongRay
    \, ,
\end{equation}
which is approximated by numerical integration
\begin{equation}
    \label{eq:vol_rendering_discrete}
    \Color(\Ray) = \sum_{i=1}^{\NumberOfSamples_r} \exp \Big[-\sum_{j=1}^{i-1} \Density_j \StepSize_j  \Big] (1 - \exp(-\Density_i\StepSize_i)) \Color(\Ray, \Direction)_i
    \, ,
\end{equation}
where $\NumberOfSamples_r$ denotes the number of samples along the ray $\Ray$ and $\StepSize_i$ is the distance between two adjacent samples.

To improve geometric surface quality in NeRF-like scene reconstructions, NeuS~\cite{wang2021neus} and VolSDF~\cite{yariv2021volsdf} propose to replace the learned density $\Density$ by a learned signed distance field $\SDF$, and then transform $\SDF$ to $\Density$ for rendering via a sigmoid-shaped map.
The formulation of NeuS optimizes an SDF $(\Color,\SDF) = \GenericNetwork(\mathbf{x}, \Direction)$ along with a global kernel size $\KernelSize$ that controls the sharpness of the implied density.
To evaluate volume rendering (\eqref{vol_rendering_discrete}) the SDF value $\SDF$ at $\Position$ is transformed to a density $\Density$ by
\begin{equation}
\label{eq:neus_remapping}
\Density = \max\left(-\frac{\frac{d\Phi_\KernelSize}{d\LengthAlongRay}(\SDF)}{\Phi_\KernelSize(\SDF)},0\right),
\qquad
  \Phi_\KernelSize(\SDF) = (1 + \exp(-\SDF / \KernelSize))^{-1},
\end{equation}
\setlength{\columnsep}{1em}
\setlength{\intextsep}{0em}
\begin{wrapfigure}{r}{0.48\columnwidth}
  \includegraphics[width=0.47\columnwidth]{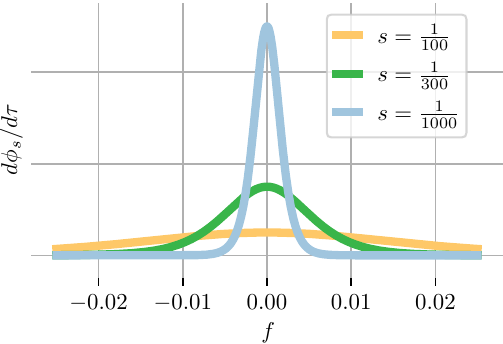}
\end{wrapfigure}
where $\SDF$ is implicitly $\SDF(\Ray(\LengthAlongRay))$ along a ray.
Intuitively, a small $\KernelSize$ results in a wide kernel with a fuzzy density, while in the limit $\lim_{\KernelSize \to 0} {d\Phi_\KernelSize}/{d\LengthAlongRay}$ approximates a sharp impulse function (see inset).
This SDF-based formulation allows for the use of an Eikonal regularizer during training, which encourages the learned $\SDF$ to be an actual distance function, resulting in a more accurate surface reconstruction.
The relevant losses are discussed in \secref{training}.

\subsection{Spatially-Varying Kernel Size}
\label{sec:extending_neus}

The NeuS SDF formulation is highly effective, yet, it relies on one \emph{global} kernel size.
In combination with the Eikonal regularization this implies a constant spread of the volume density across the whole scene.
However, a one-size-fits-all approach does not adapt well to
scenes that contain a mixture of ``sharp'' surfaces (\eg{} furniture or cars) and ``fuzzy'' volumetric regions (\eg{} hair or grass).

\begin{figure}
    \centering
    \includegraphics{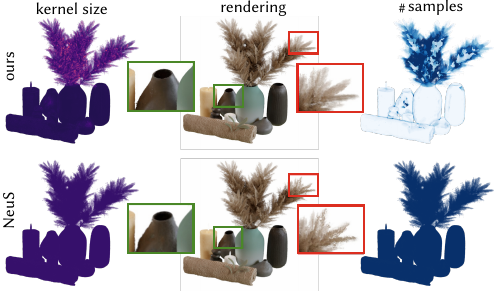}
    \caption{
    The original NeuS~\cite{wang2021neus} formulation uses a single global kernel size $\KernelSize$.
    On complex scenes with varying content, the global $\KernelSize$ value converges to an average which is too small for volumetric parts~\protect\tikz \protect\node [rectangle,draw, thick, red!90] at (2,-2) {};~, and too large for sharp surfaces~\protect\tikz \protect\node [rectangle,draw, thick, green_zoomin] at (2,-2) {};~.
    Instead, our locally varying kernel size adapts to the scene, in-turn allowing us to reduce the number of samples to a single sample~\tikzcircle[black,fill=white]{2.5pt} for sharp surfaces and up to 32 samples~\tikzcircle[black,fill=nsteps_blue]{2.5pt} for the fern (\emph{top right}). NeuS uses a constant 384 samples per pixel (\emph{bottom right}).}
    \label{fig:variable_kernel_size}
\end{figure}

Our first contribution is to augment the NeuS formulation with a spatially-varying, \emph{locally} learned kernel size $\KernelSize$ as an additional neural output that is dependent on the input 3D position $\Position$. The extended network becomes $(\Color,\SDF, \KernelSize) = \GenericNetwork(\Position, \Direction)$ (see the implementation details in \secref{architectures}).
During training, we additionally include a regularizer that promotes the smoothness of the kernel size field (\secref{training}).
This neural field can still be fit from only color image supervision, and the resulting spatially-varying kernel size automatically adapts to the sharpness of the scene content (\figref{variable_kernel_size}).
This enhanced representation is independently valuable, improving reconstruction quality in difficult scenes, but importantly it will serve to guide our explicit shell extraction in \secref{shell_extraction}, which greatly accelerates rendering.

\begin{figure}
    \centering
    \includegraphics{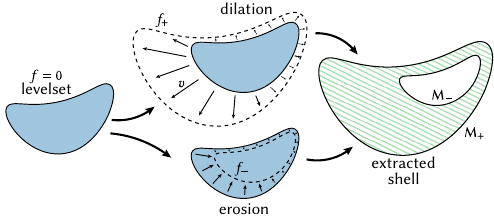}
    \caption{
        After fitting an initial SDF and spatially varying kernel, we apply level set flows to extract an adaptive shell via dilation and erosion.
        For illustrative purposes, the adaptive shell is enlarged; in practice it very tightly encloses sharp surfaces.
        \label{fig:levelset_band}
    }
\end{figure}

\subsection{Extracting an Explicit Shell}
\label{sec:shell_extraction}

The adaptive shell delimits the region of space which contributes significantly to the rendered appearance, and is represented by two explicit triangle meshes.
When $\KernelSize$ is large the shell is thick, corresponding to volumetric scene content, and when $\KernelSize$ is small the shell is thin, corresponding to surfaces.
After the implicit fields $\KernelSize$ and $\SDF$ have been fit as described in \secref{extending_neus}, we extract this adaptive shell once as a post-process.

In \eqref{neus_remapping} the magnitude of the quantity $\SDF / \KernelSize$ in the sigmoid exponent determines the rendering contribution along a ray (see the inset figure in \secref{preliminaries}).
It is tempting to simply extract a band where $|\SDF / \KernelSize| < \eta$ for some $\eta$ as the region that makes a significant contribution to the rendering.
However, the learned functions quickly become noisy away from the $\SDF = 0$ level set, and cannot be sufficiently regularized without destroying fine details.
Our solution is to separately extract an inner boundary as an erosion of the $\SDF=0$ level set, and an outer boundary as its dilation (\figref{levelset_band}), both implemented via a regularized, constrained level set evolution tailored to the task.

In detail, we first sample the fields $\SDF$ and $s$ at the vertices of a regular grid.
We then apply a level set evolution to $\SDF$, producing a new eroded field $\ErodedField$, and extract the $\ErodedField = 0$ level set as the inner shell boundary via marching cubes.
A separate, similar evolution yields the dilated field $\DilatedField$, and the $\DilatedField = 0$ level set forms the outer shell boundary.
We define both level sets separately: the dilated outer surface should be smooth to avoid visible boundary artifacts, while the eroded inner surface needs not be smooth, but must only exclude regions which certainly do not contribute to the rendered appearance.

Recall that a basic level set evolution of a field $a$ is given by ${\partial a}/{\partial t} = -\left|\nabla a\right| \LevelsetVelocity $, where $\LevelsetVelocity$ is the desired scalar outward-normal velocity of the level set.
Our constrained, regularized flow on $\SDF$ is then
\begin{equation}
\label{eq:levelset_flow}
  \frac{\partial \SDF}{\partial t} =
    -\left|\nabla \SDF\right| \bigg(
      \LevelsetVelocity(\SDF_0, s) +
      \lambda_\textrm{curv} \nabla \cdot \dfrac{\nabla \SDF}{|\nabla \SDF|}
      \bigg)
    \omega(\SDF),
\end{equation}
\setlength{\columnsep}{1em}
\setlength{\intextsep}{0em}
\begin{wrapfigure}{r}{80pt}
  \includegraphics{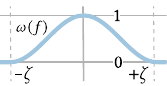}
\end{wrapfigure}
where $\SDF_0$ here denotes the initial learned SDF, the divergence term is a curvature smoothness regularizer with weight $\lambda_\textrm{curv}$.
The soft falloff $\omega$ (see inset) limits the flow to a window around the level set:
\begin{equation}
\label{eq:falloff}
  \omega(\SDF) = \tfrac{1}{2}\big(1 + \cos(\pi\ \textrm{clamp}(\SDF / \zeta, -1., 1.)\big),
\end{equation}
with window width $\zeta$.
To dilate the level set, the velocity is chosen to fill all regions with density $\Density > \DensityMin$ for a ray incoming in the normal direction
\begin{equation}
\label{eq:dilation}
  \LevelsetVelocity_\textrm{dilate}(\SDF_0,\KernelSize) =
    \begin{cases}
      \DensityDilateVelCoef \Density(\SDF_0,\KernelSize)  & \Density(\SDF_0,\KernelSize) > \DensityMin \\
      0 & \Density(\SDF_0,\KernelSize) \leq \DensityMin
    \end{cases},
\end{equation}
with $\DensityDilateVelCoef$ as a scaling coefficient.
We use $\zeta = 0.1$, and $\lambda_\textrm{curv} = 0.01$.
To erode the level set, the velocity is inversely-proportional to density, so that the shell expands inward quickly for low density regions and slowly for high density regions
\begin{equation}
\label{eq:erosion}
  \LevelsetVelocity_\textrm{erode}(\SDF_0,\KernelSize) = \min\big(\ErodeVelocityMax,\DensityErodeVelCoef \dfrac{1}{\Density(\SDF_0,\KernelSize)}\big),
\end{equation}
where here we use $\zeta = 0.05$, and $\lambda_\textrm{curv} = 0$.
These velocities lead to a short-distance flow, and thus a narrow shell where $\KernelSize$ is small and the content is surface-like.
They lead to a long-distance flow and hence a wide shell where $\KernelSize$ is large and the content is volume-like.

\begin{figure}
    \centering
    \includegraphics{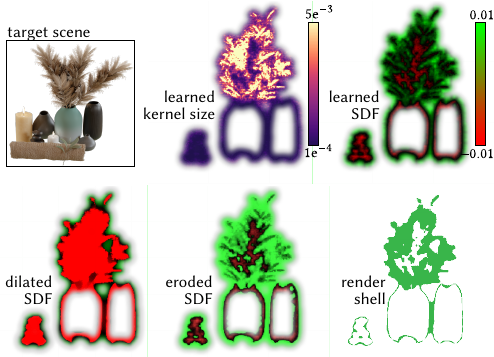}
    \caption{
        Given kernel size $\KernelSize$ and SDF $\SDF$ learned from \secref{extending_neus} (\emph{top}), we apply a erosion and dilation flows to $\SDF$  (\emph{bottom middle and left}) to extract a narrow shell in which we efficiently render (\emph{bottom right}).
        Here, we visualize each quantity on a 2D slice through a scene.
        For clarity, we show the fields only nearby the adaptive shell that is ultimately rendered.
        \label{fig:levelset_slices}
    }
\end{figure}

We compute the dilated field $\DilatedField$ and eroded field $\ErodedField$ respectively by forward-Euler integrating this flow on the grid for $50$ steps of integration, computing derivatives via spatial finite differences.
We do not find numerical redistancing to be necessary.
Finally, we clamp the results $\ErodedField \gets$ $\max(\SDF_0, \ErodedField)$ and $\DilatedField \gets$ $\min(\SDF_0, \DilatedField)$, to ensure that the eroded field only shrinks the level set, and the dilated flow only grows the level set.
The $\DilatedField = 0$ and $\ErodedField = 0$ level sets are extracted via marching cubes as the outer and inner shell boundary meshes $\OuterMesh$ and $\InnerMesh$, respectively.
\figref{levelset_slices} visualizes the resulting fields.
Further details are provided in Procedure~\ref{proc:levelset} and~\ref{proc:shellextraction} of the Appendix.

\subsection{Narrow-Band Rendering}
\label{sec:narrow_band_rendering}

The extracted adaptive shell serves as an auxiliary acceleration data structure to guide the sampling of points along a ray (\eqref{vol_rendering_discrete}), enabling us to efficiently skip empty space and sample points only where necessary for high perceptual quality.
For each ray, we use hardware-accelerated ray tracing to efficiently enumerate the ordered intervals defined by the intersection of the ray and the
\setlength{\columnsep}{1em}
\setlength{\intextsep}{0em}
\begin{wrapfigure}{r}{100pt}
  \includegraphics{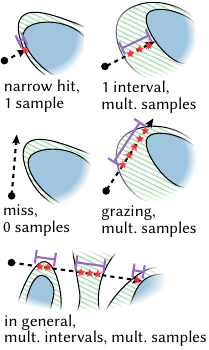}
\end{wrapfigure}
adaptive shell. Within each interval we query equally-spaced samples.
Our renderer does not require any dynamic adaptive sampling or sample-dependent termination criteria, which facilitates high-performance parallel evaluation.

In detail, we first build ray tracing acceleration data structures for both the outer mesh $\OuterMesh$ and inner mesh $\InnerMesh$
We then cast each ray against the meshes, yielding a series of intersections where the ray enters or exits the mesh, partitioning the ray into zero or more intervals contained in the shell (see inset).
For each interval with width $\IntervalWidth$, a target inter-sample spacing $\IntervalSampleSpacing$, and a single-sample threshold $\IntervalSingleSampleWidth$, we compute the number of samples as $\mathrm{ceil}(\mathrm{max}(\IntervalWidth-\IntervalSingleSampleWidth, 0) / \IntervalSampleSpacing)+1$.
We cap the maximum number of samples to $\NIntervalSamplesMax$, and equidistantly sample the interval.
Note that if the interval has $\IntervalWidth < \IntervalSingleSampleWidth$, a single sample is taken at the center of the interval.
When an interval ends because the ray hits the inner mesh $\InnerMesh$, we do not process any subsequent samples, as this represents the interior of a solid object.
Otherwise, we process intervals until the ray exits the scene or we hit a maximum cap, accumulating the contributions as in \eqref{vol_rendering_discrete}.

Note that this procedure can be implemented by first generating all samples within all intervals, and then performing a single batched MLP inference pass, which improves throughput.
For surfaces, our narrow-band sampling often amounts to just a hardware-accelerated ray tracing, followed by a single network evaluation, while for fuzzy regions it densely samples only where necessary---in either case, performance is greatly accelerated (\tabref{perf_comparison}).
More algorithmic details are included in Procedure~\ref{proc:sampling} of the Appendix.

\begin{figure}
    \centering
    \includegraphics{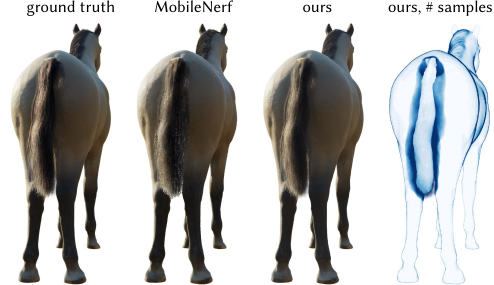}
    \caption{Pure surface-based representations struggle to represent fuzzy surfaces such as the tail. On the other hand, our method adapts the narrow shell to the local complexity of the scene, using a single sample~\tikzcircle[black,fill=white]{2.5pt} for the sharp skin surface and up to 16 samples~\tikzcircle[black,fill=nsteps_blue]{2.5pt} for the tail.}
    \label{fig:horse}
\end{figure}

\subsection{Losses and Training}
\label{sec:training}

We optimize the parameters of our representation in two stages. In the first stage, we use the fully volumetric formulation described in Sections~\ref{sec:preliminaries} and \ref{sec:extending_neus} and minimize the following objective
\begin{equation}
\label{eq:loss}
\Loss = \Loss_\Color + \LossWeight_e\Loss_e + \LossWeight_\KernelSize\Loss_\KernelSize  + \LossWeight_\NormalVector\Loss_\NormalVector
\end{equation}
with the weights $\LossWeight_\Color = 1$, $\LossWeight_e = 0.1$, $\LossWeight_\NormalVector =  0.1$, and $\LossWeight_\KernelSize = 0.01$ for all experiments.
Here $\Loss_\Color$ is the standard pixel-wise color loss against calibrated ground-truth images
\begin{equation}
\label{eq:loss_color}
\Loss_\Color = \frac{1}{|\SetOfRays|}\sum_{\Ray \in \SetOfRays }|\Color(\Ray) - \Color_\textrm{gt}(\Ray)|
\end{equation}
and $\Loss_e$ is the Eikonal regularizer as in \cite{wang2021neus}
\begin{equation}
\Loss_e = \frac{1}{|\SetOfSamples|} \sum_{\Position \in \SetOfSamples}\left(||\nabla f(\Position)||_2 - 1\right)^2,
\end{equation}
where $\SetOfRays$ and $\SetOfSamples$ denote the set of rays and samples along the rays, respectively.
$\nabla f(\mathbf{x})$ can be obtained either analytically~\cite{wang2021neus, yariv2021volsdf, wang2022neus2} or through finite differences~\cite{Li2023Neuralangelo, wang2023fegr}.
We use the latter approach.
The loss $\Loss_\KernelSize$ regularizes the spatially varying kernel size introduced in our formulation for smoothness
\begin{equation}
\Loss_\KernelSize = \frac{1}{\SetOfSamples} \sum_{\Position \in \SetOfSamples} || \log\big[\KernelSize(\Position) \big] -  \log\big[ \KernelSize(\Position + \mathcal{N}(0, \NoiseSTD^2)) \big] ||_2,
\end{equation}
where $\mathcal{N}(0, \NoiseSTD^2)$ denotes samples from the normal distribution with standard deviation $\NoiseSTD$.
Lastly, we incorporate the loss 
\begin{equation}
\Loss_\NormalVector = \frac{1}{|\SetOfSamples|}\big|\big|\NormalVector(\Position) - \frac{\nabla f(\Position)}{||\nabla f(\Position)||_2} \big|\big|_2,
\end{equation}
internal to our network architecture (\secref{architectures}).
Like NeuS, we will leverage geometric normals as an input to a shading subnetwork, but we find that predicting these normals internally improves inference performance \vs{} gradient evaluation.
$\Loss_\NormalVector$ serves to train these predicted normals to remain approximately faithful to ones obtained through the finite differences of the underlying SDF field $\nabla f(\mathbf{x})$~\cite{wang2023fegr, Li2023Neuralangelo}.

After the implicit field has been fit, we extract the adaptive shell as in \secref{shell_extraction}.
While initial training requires dense sampling along rays, our explicit shell now allows narrow-band rendering to concentrate samples only in significant regions.
We therefore fine-tune the representation within the narrow band, now with only $\Loss_\Color$---it is no longer necessary to encourage a geometrically-nice representation, as we have already extracted the shell and restricted the sampling to a small band around the scene content. Disabling regularization enables the network to devote its whole capacity to fit the visual appearance, which leads to improved visual fidelity (\tabref{shelly_ablation}).
In Procedure~\ref{proc:training} of the Appendix, we also present the training pipeline with algorithm details.

\section{Experiments}
\label{sec:experiments}
In this section, we provide low-level details of our implementation and evaluate our method in terms of rendering quality and efficiency on four data sets that range from synthetic object-level \ShellyDataset{}, \NeRFDataset{}~\cite{mildenhall2020nerf} and tabletop \DTUDataset~\cite{jensen2014large} data, to real-world large outbound scenes \MipNerfDataset~\cite{barron2022mipnerf360}.
For comparisons, we treat Instant NGP~\cite{mueller2022ingp} as our volumetric baseline, due to its balance between high fidelity and efficiency. In addition, we compare to prior methods that were optimized either for fidelity \cite{yuan2022nerf, wang2021neus, barron2021mipnerf, barron2022mipnerf360} or rendering efficiency~\cite{chen2022mobilenerf, yariv2023bakedsdf,guo2023vmesh}.

When running NeRF~\cite{mildenhall2020nerf} and Mip-NeRF~\cite{barron2021mipnerf} on \DTUDataset{} and \ShellyDataset, we use the implementation from Nerfstudio~\cite{nerfstudio}. For other methods and experiment settings, we use their official implementations.

\subsection{Architecture Details}
\label{sec:architectures}

Following the state of the art in neural volumetric rendering, we represent our neural field as a combination of a feature field and a small (decoder) neural network. Specifically, we use a multi-resolution hash encoding~\cite{mueller2022ingp} $\PosEncoding(\cdot)$ with 14 levels, where each level is represented by a hash-table with $2^{22}$ two-dimensional features. The voxel grid resolution of our feature field grows from $16^3 \to 4096^3$ for \ShellyDataset{} and \NeRFDataset{}, and from $16^3 \to 8192^3$ for the other data sets. The features at each level are obtained through tri-linear interpolation before being concatenated to form the feature embedding $\PosEncoding(\cdot) \in \mathbb{R}^{28}$. This is further concatenated with the sample coordinates and input to the geometry network $(\SDF, 1/\KernelSize, \GeoFeature, \NormalVector) = \GeometryNetwork([\PosEncoding(\Position),\Position] )$ which is an MLP with a single hidden layer and $31 \to 64 \to 31$ dimensions. Apart from the SDF value $\SDF$ and kernel size $\KernelSize$, $\GeometryNetwork$ also outputs a geometry latent feature $\GeoFeature \in \mathbb{R}^{26}$ and the normal vector $\NormalVector \in \mathbb{R}^{3}$ which are combined with $\Position$ and the encoded view direction $\Direction$ as input to the radiance network $\Color = \RadianceNetwork([\DirEncoding(\Direction), \GeoFeature, \NormalVector, \mathbf{x}])$ that predicts the emitted color. Here, $\RadianceNetwork$ is an MLP with two hidden layers and dimensions $48 \to 64 \to 64 \to 3$. To reduce the computational cost, we directly predict the normal vector $\NormalVector$ with an MLP rather than computing it as the gradient of the underlying SDF field as done in NeuS~\cite{wang2021neus}. Finally, to encode the input direction $\Direction$, we use the spherical harmonic basis up to degree 4, such that $\DirEncoding(\Direction) \in \mathbb{R}^{16}$. The dimensions of all layers in both networks and the feature field were selected for high throughput on modern GPU devices.

\subsection{Implementation}
\label{sec:implementation}
The training stage of our method is implemented in PyTorch~\cite{paszke2017pytorch}, while the inference stage is implemented in Dr.Jit~\cite{Jakob2022DrJit}.
To achieve real-time inference rates, we rely on the automatic kernel fusion performed by Dr.Jit as well as GPU-accelerated ray-mesh intersection provided by OptiX~\cite{parker2010optix}.
While the inference pass is implemented with high-level Python code, the asynchronous execution of large fused kernels hides virtually all of the interpreter's overhead.
Combined with the algorithmic improvements described above, we achieve frame rates from 40~fps (25~$\nicefrac{\text{ms}}{\text{frame}}$) on complex outdoor scenes to 300~fps (3.33~$\nicefrac{\text{ms}}{\text{frame}}$) on object-level scenes, at 1080p resolution on a single RTX 4090 GPU.
A performance comparison to Instant NGP~\cite{mueller2022ingp} on four data sets is given in \tabref{perf_comparison}.
Note that in this work, we focused on inference performance only, and have not yet applied these performance optimizations to the training procedure.
Detailed pseudo-code is given in Procedures~\ref{proc:levelset},~\ref{proc:shellextraction},~\ref{proc:sampling} and \ref{proc:training} of the Appendix.

\begin{table*}[thbp]
\setlength{\tabcolsep}{2pt}
\centering
\small
\caption{
    Performance comparisons on all \revReplaced{three}{four} data sets, measured at 1080p without GUI overhead using an RTX~4090 GPU.
    Our adaptive sample placement and mesh-based empty-space skipping technique allows us to outperform Instant NGP without compromising visual fidelity.
    \emph{Note that Instant NGP's performance on the DTU data set was hindered by a large number of background samples, and is therefore not necessarily indicative of a real use case: the user may specify a tighter scene bounding box to focus the samples on the main scene contents.}
}
\label{tab:perf_comparison}
\begin{tabular}{cc|ccc|ccc}
\toprule
& & \multicolumn{3}{c|}{Ours} & \multicolumn{3}{c}{Instant NGP~\cite{mueller2022ingp}} \\
& & Sample count~$\downarrow$ & \nicefrac{\text{ms}}{\text{frame}}~$\downarrow$ & FPS~$\uparrow$ & Sample count~$\downarrow$ & \nicefrac{\text{ms}}{\text{frame}}~$\downarrow$ & FPS~$\uparrow$ \\
\midrule
& \ShellyDataset          &  2.07 &  \revAdded{3.81} & \revAdded{262.69} & \revAdded{2.89}  &  \revAdded{11.74} & \revAdded{85.16} \\
& \DTUDataset             &  5.11 &  \revAdded{6.37} & \revAdded{157.00} & 56.10 & \revAdded{123.31} &  \revAdded{8.10} \\
& \revAdded{\NeRFDataset} & \revAdded{1.98} & \revAdded{3.56} & \revAdded{280.68} & \revAdded{3.19} & \revAdded{14.20} & \revAdded{70.40} \\
& \MipNerfDataset         & 17.05 & \revAdded{27.64} & \revAdded{36.18} & 45.62  &  \revAdded{93.57} & \revAdded{10.69} \\
\bottomrule
\end{tabular}
\end{table*}

\subsection{Evaluation Metrics}
In order to evaluate the rendering quality, we report the commonly used peak signal-to-noise ratio (\PSNR), learned perceptual image patch similarity (\LPIPS), and structural similarity (\SSIM) metrics. Unfortunately, evaluating the efficiency of the methods is less straightforward as the complexity of the method is often intertwined with the selected hardware and low-level implementation details. Indeed, reporting only the number of frames-per-second (FPS) or the time needed to render a single frame may paint an incomplete picture. We therefore additionally report the number of samples along the ray that are required to render each pixel. While the number of samples along the ray also does not tell the whole story as the per-sample evaluation can have different computational complexity, combining all metrics provides a good assessment of the computational complexity of the individual methods.

\begin{table*}[!thbp]
\setlength{\tabcolsep}{6pt}
\centering
\small
\caption{Quantitative results on \ShellyDataset{} data set, \DTUDataset{} data set \revAdded{and \NeRFDataset{} data set}. We report PSNR, LPIPS and SSIM. Our method achieves better results across all metrics \revAdded{on \ShellyDataset{} and \DTUDataset{} and comparable results on \NeRFDataset}. \emph{Real-time} denotes methods that achieve >30FPS at 1080p. \revAdded{On \ShellyDataset{} and \DTUDataset, we run NeRF and Mip-NeRF with Nerfstudio~\cite{nerfstudio}, and use official implementation for other methods. Baselines of \NeRFDataset{} are from the original papers.} Detailed results for each object/scene are provided in the Supplement.
}
\label{tab:main_table}
\begin{tabular}{cc|lll|lll|lll}
\toprule
& & \multicolumn{3}{c|}{\ShellyDataset} & \multicolumn{3}{c|}{\DTUDataset} & \multicolumn{3}{c}{\revAdded{\NeRFDataset}}\\
& & PSNR~$\uparrow$ & SSIM~$\uparrow$ & LPIPS~$\downarrow$ & PSNR~$\uparrow$ & SSIM~$\uparrow$ & LPIPS~$\downarrow$ & \revAdded{PSNR~$\uparrow$} & \revAdded{SSIM~$\uparrow$} & \revAdded{LPIPS~$\downarrow$} \\
\midrule
\multirow{4}{*}{\rotatebox[origin=c]{90}{\emph{offline}}}
& NeRF~\cite{mildenhall2020nerf} & 31.27 & 0.893 & 0.157 & 28.51	& 0.894 & 0.183	& \revAdded{31.01} & \revAdded{0.947} & \revAdded{0.081} \\
& NeuS~\cite{wang2021neus}   & 29.98 & 0.893 & 0.158 & 28.92 &	0.913~\tikzcircle[silver,fill=silver]{2pt} &0.168 & \revAdded{/} & \revAdded{/} & \revAdded{/} \\
& Mip-NeRF~\cite{barron2021mipnerf} & 32.59 & 0.899 & 0.148 & 28.90&	0.900 & 0.179 & \revAdded{33.09}~\tikzcircle[silver,fill=silver]{2pt} & \revAdded{0.961}~\tikzcircle[silver,fill=silver]{2pt} & \revAdded{0.043}~\tikzcircle[gold,fill=gold]{2pt} \\
& Ours (full ray) & 34.26~\tikzcircle[silver,fill=silver]{2pt} & 0.932~\tikzcircle[silver,fill=silver]{2pt} & 0.104~\tikzcircle[silver,fill=silver]{2pt} & 33.51~\tikzcircle[gold,fill=gold]{2pt}& 	0.901 & 0.081~\tikzcircle[silver,fill=silver]{2pt} & \revAdded{32.51}~\tikzcircle[bronze,fill=bronze]{2pt} & \revAdded{0.962}~\tikzcircle[gold,fill=gold]{2pt} & \revAdded{0.048}~\tikzcircle[silver,fill=silver]{2pt} \\
\midrule
\midrule
\multirow{4}{*}{\rotatebox[origin=c]{90}{\emph{real-time}}}
& I-NGP~\cite{mueller2022ingp} & 33.22~\tikzcircle[bronze,fill=bronze]{2pt}  & 0.922~\tikzcircle[bronze,fill=bronze]{2pt} & 0.125~\tikzcircle[bronze,fill=bronze]{2pt} & 31.37~\tikzcircle[bronze,fill=bronze]{2pt} & 0.932~\tikzcircle[silver,fill=silver]{2pt} & 	0.139~\tikzcircle[bronze,fill=bronze]{2pt} & \revAdded{33.18}~\tikzcircle[gold,fill=gold]{2pt} & \revAdded{/} & \revAdded{/}  \\
& MobileNeRF~\cite{chen2022mobilenerf} & 31.62 & 0.911 & 0.129 & / & / & / & \revAdded{30.90} & \revAdded{0.947} & \revAdded{0.062}  \\
& \revAdded{VMesh~\cite{guo2023vmesh}} & \revAdded{/} & \revAdded{/} & \revAdded{/} & \revAdded{/} & \revAdded{/} & \revAdded{/} & \revAdded{30.70} & \revAdded{0.947} & \revAdded{0.060} \\
& Ours & 36.02~\tikzcircle[gold,fill=gold]{2pt} & 0.954~\tikzcircle[gold,fill=gold]{2pt} & 0.079~\tikzcircle[gold,fill=gold]{2pt} & 33.37~\tikzcircle[silver,fill=silver]{2pt}	&	0.964~\tikzcircle[gold,fill=gold]{2pt} & 0.077~\tikzcircle[gold,fill=gold]{2pt} & \revAdded{31.84} & \revAdded{0.957}~\tikzcircle[bronze,fill=bronze]{2pt} & \revAdded{0.056}~\tikzcircle[bronze,fill=bronze]{2pt} \\
\bottomrule
\end{tabular}
\end{table*}

\subsection{\ShellyDataset{} Data Set}
\label{sec:shelly}
The \NeRFDataset{} data set that was introduced in \cite{mildenhall2020nerf} is still one of the most widely used data sets to evaluate novel-view synthesis methods.
Yet, it mainly consists of objects with sharp surfaces that can be well-represented by surface rendering methods~\cite{Munkberg2022nvdiffrec}, and thus does not represent the challenge of general scene reconstruction.
This motivated us to introduce a new synthetic data set, which we name {\ShellyDataset}.
It covers a wider variety of appearance including fuzzy surfaces such as hair, fur, and foliage.
{\ShellyDataset} contains six object-level scenes: \textsc{Khady}, \textsc{Pug}, \textsc{Kitty}, \textsc{Horse}, \textsc{Fernvase} and \textsc{Woolly}. For each scene, we have rendered 128 training and 32 test views from random camera positions distributed on a sphere with a fixed radius.
We are grateful to the original artists of these models: \textsc{jhon maycon}, Pierre-Louis Baril, \textsc{abdoubouam}, \textsc{ckat609}, the BlenderKit team, and \textsc{textures.xyz}.

\begin{figure*}
    \centering
    \includegraphics[height=\textheight]{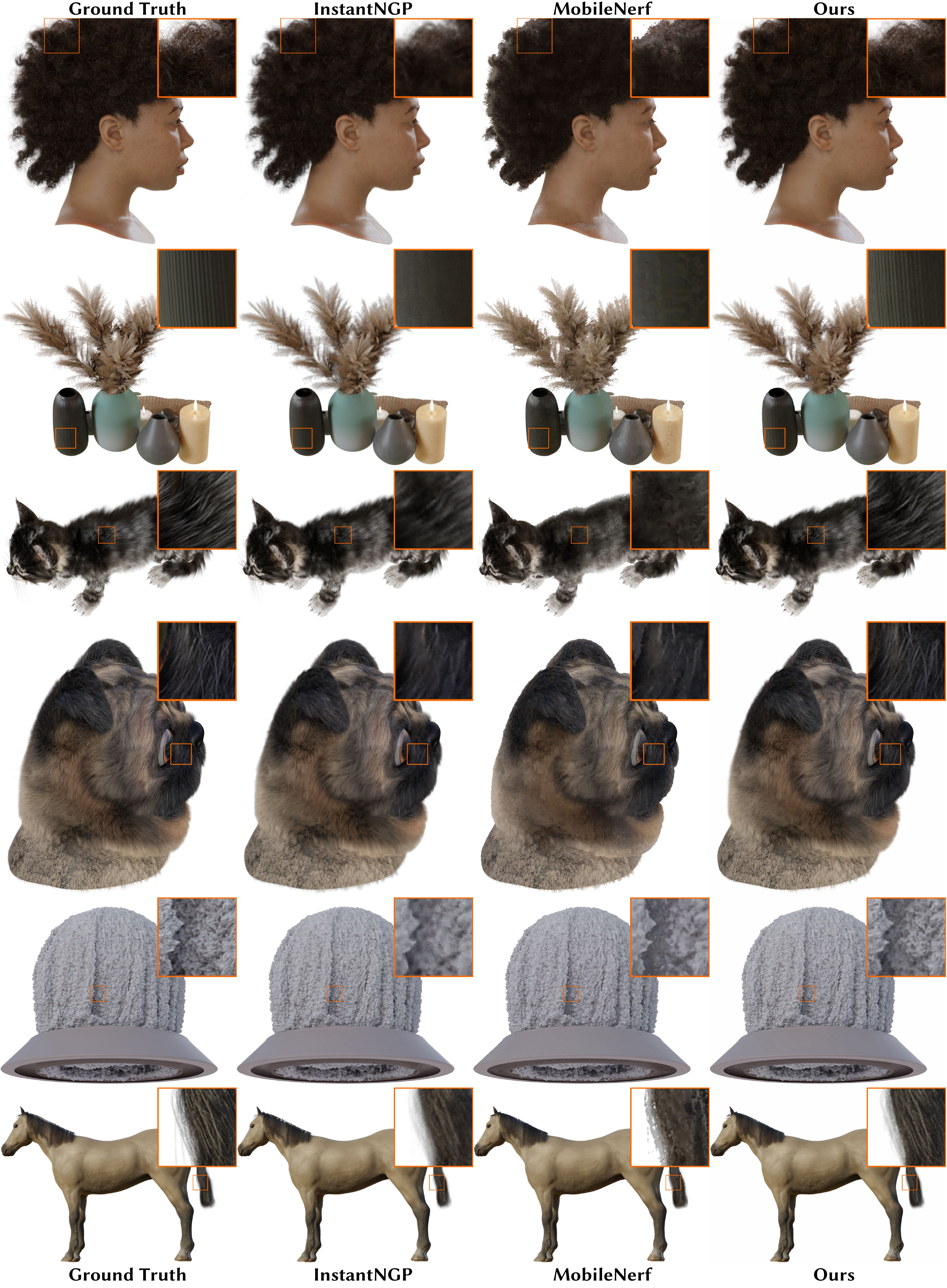}
    \vspace*{-1em}
    \caption{
        A gallery of results on the test-views of our \ShellyDataset{} data set.
        \label{fig:gallery_shelly_compress}
    }
\end{figure*}

\tabref{main_table} shows quantitative results, while the novel views are qualitatively compared in
\figref{gallery_shelly_compress}. Our method significantly outperforms prior methods across all quality metrics achieving more than 2dB higher PSNR than Instant NGP. \figref{gallery_shelly_compress} demonstrates that surface-based rendering methods (MobileNerf) struggle to represent fuzzy surfaces. On the other hand, our method aligns its representation to the complexity of the scene. For example, \figref{horse} shows that our method represents the skin of the horse as a sharp surface, while using a wider kernel for its tail, which benefits from volumetric rendering.

\subsection{\DTUDataset{} Data Set}
\label{sec:dtu}
We consider 15 tabletop scenes from the \DTUDataset{} data set~\cite{jensen2014large}.  These scenes were captured by a robot-held monocular RGB camera, and are commonly used to evaluate implicit surface representations. We follow prior work~\cite{wang2021neus, yariv2021volsdf} and task the methods to represent the full scene, but evaluate the performance only within the provided object masks.

\begin{figure*}
    \centering
    \includegraphics[width=1\textwidth]{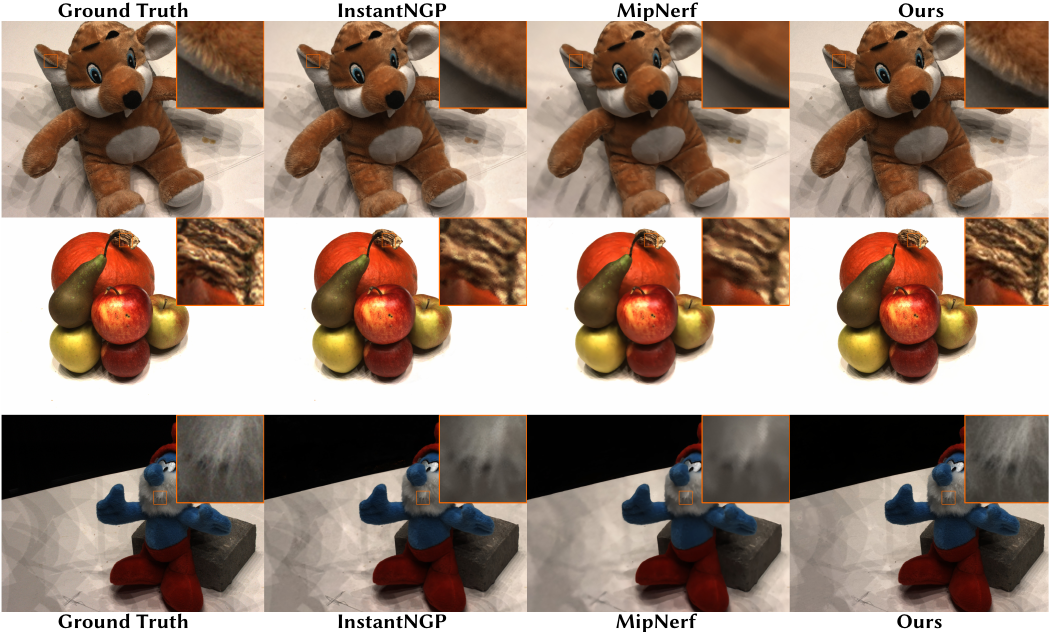}
    \vspace*{-1em}
    \caption{
        A gallery of results on the \DTUDataset{} data set.
        \label{fig:gallery_dtu}
    }
\end{figure*}

\tabref{main_table} depicts that our method outperforms all baselines across all evaluation metrics. Qualitative results are provided in \figref{gallery_dtu}.
Different from the \ShellyDataset~ data set, the performance of \emph{Ours} on the \DTUDataset{} data set is slightly lower than that of \emph{Ours (full ray)} in terms of PSNR. We hypothesize that this is due to the distribution of the camera poses that observe the scene only from a single direction. This hinders constraining the neural field and hence also the adaptive shell extraction. The same reason also contributes to a significant increase in the sample count for Instant NGP (see \tabref{perf_comparison}).

\subsection{\NeRFDataset{} Data Set}
\label{sec:nerf-synthetic}
The \NeRFDataset{} data set introduced in \cite{mildenhall2020nerf} contains 8 synthetic objects rendered in Blender and is widely adopted to evaluate the quality of novel view synthesis methods.

As shown in \tabref{main_table}, our method can achieve comparable quality to the state of the art methods Mip-NeRF~\cite{barron2021mipnerf} and I-NGP~\cite{mueller2022ingp}, but with a much faster runtime performance (\tabref{perf_comparison}).
Our method also achieves better image quality compared to recent works optimized for rendering efficiency~\cite{chen2022mobilenerf,guo2023vmesh}\footnote{Note that VMesh~\cite{guo2023vmesh} also optimizes for disk storage, which is an orthogonal direction to this paper.}.

\subsection{\MipNerfDataset~ Data Set}
\label{sec:mipnerf_experiments}
The MipNeRF-360 data set~\cite{barron2022mipnerf360} is a challenging real-world data set that contains large indoor and outdoor scenes captured from $360^\circ$ camera views\footnote{In our evaluation, we exclude the two scenes with license issues: \textit{Flowers} and \textit{Treehill}.}. The scenes feature a complex central object accompanied by a highly detailed background. To better represent the background, we follow \cite{yariv2023bakedsdf} and extend our method with the scene contraction proposed in \cite{barron2022mipnerf360} (more details are provided in the supplemental document).

\begin{figure*}
    \centering
    \includegraphics[width=1\textwidth]{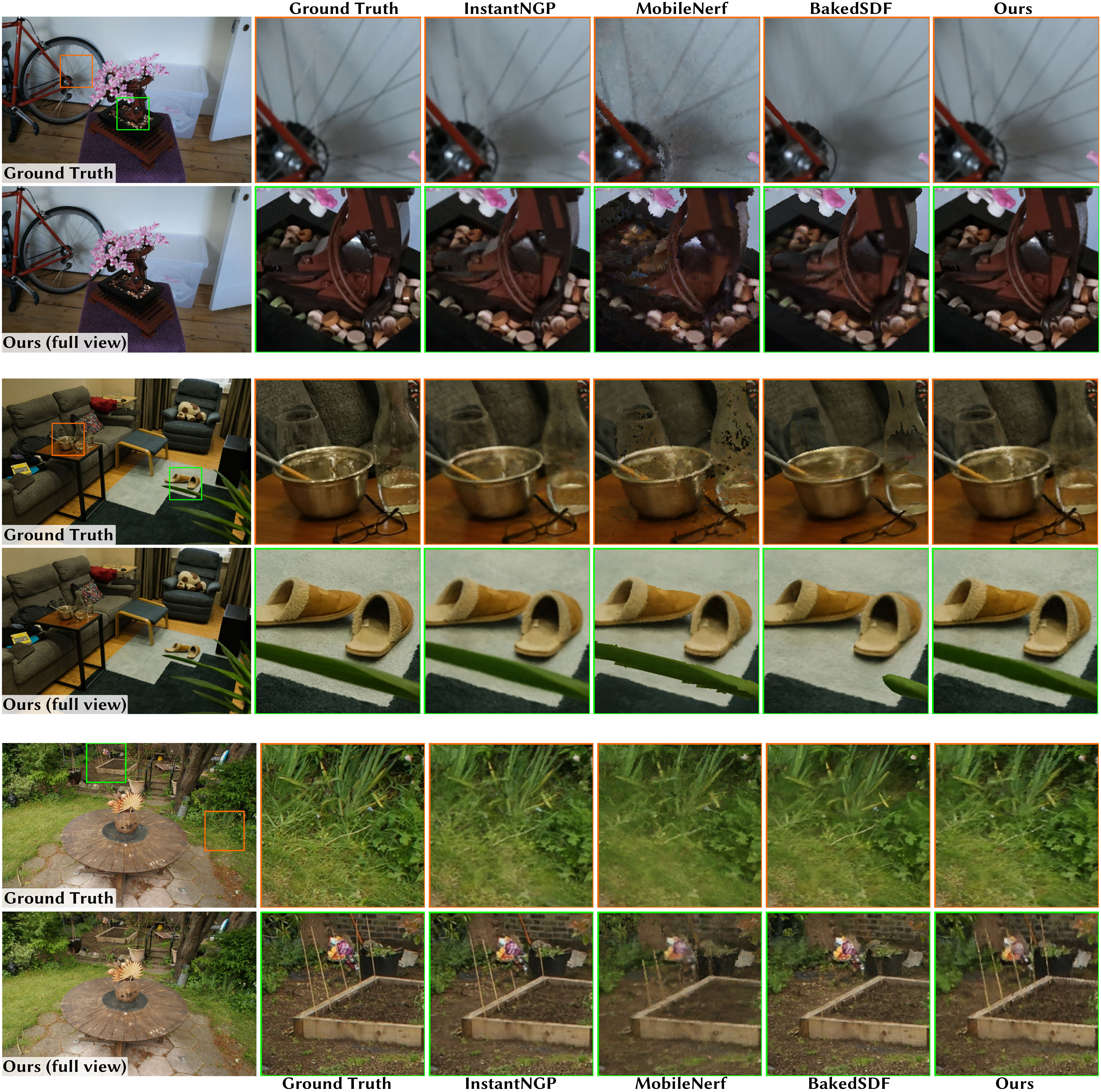}
    \vspace*{-1em}
    \caption{
        A gallery of results on the test-views of the MipNerf360 data set.
        \label{fig:gallery_mipnerf360}
    }
\end{figure*}

\tabref{outdoor_scenes} provides the quantitative results and the qualitative comparison is depicted in \figref{gallery_mipnerf360}. Our method achieves comparable performance to other \textit{interactive} methods. Directly compared to I-NGP, our proposed narrow-band formulation can reduce the number of samples by a factor of three, resulting in fivetimes higher average frame rates at comparable rendering quality.
We note that on this data set, performance and quality depend significantly on the background, which our approach is not specialized to handle.

\subsection{Performance Evaluation}
\label{sec:performance_evaluation}
We compare the performance of our method to the most efficient volumetric baseline, Instant NGP~\cite{mueller2022ingp}, in \tabref{perf_comparison}. To ensure a fair comparison, we render the same test views for both methods at 1080p resolution and remove the GUI overhead. The comparison was run on a single RTX~4090 GPU. Our narrow-band rendering formulation can efficiently reduce the number of samples along the ray (up to 10 times) which results in significantly reduced inference time per frame.
On the challenging outbound 360 scenes, our method already runs at real-time rates.
Yet, additional speed-ups could be achieved by further studying the interaction of our adaptive sample placement with the spatial remapping employed in these scenes.

\begin{table}[t!]
\setlength{\tabcolsep}{2pt}
\centering
\small
\caption{Quantitative results on the \MipNerfDataset~ data set. We report the PSNR, LPIPS and SSIM results for each object and compare them to baselines. Our method achieves a performance comparable to the baselines while being significantly faster during inference\revAdded{~(see \tabref{perf_comparison})}.
\revAdded{In our comparison, we exclude the two scenes with license issues: Flowers, Treehill.}
}
\label{tab:outdoor_scenes}
\vspace{-1em}
\resizebox{1\columnwidth}{!}{
\begin{tabular}{cc|lll|lll}
\toprule
  & & \multicolumn{3}{c}{Outdoor scenes} & \multicolumn{3}{c}{Indoor scenes} \\
  & & PSNR~$\uparrow$ & SSIM~$\uparrow$ & LPIPS~$\downarrow$ & PSNR~$\uparrow$ & SSIM~$\uparrow$ & LPIPS~$\downarrow$ \\
  \midrule
  \multirow{4}{*}{\rotatebox[origin=c]{90}{\emph{offline}}} & NeRF~\cite{mildenhall2020nerf} & 22.20 & 0.485 & 0.501 & 26.84 & 0.790 & 0.370 \\
  & Mip-NeRF~\cite{barron2021mipnerf} & 22.02 & 0.505 & 0.484 & 26.98 & 0.798 & 0.361  \\
  & Mip-NeRF 360~\cite{barron2022mipnerf360} & 25.92~\tikzcircle[gold,fill=gold]{2pt} & 0.747~\tikzcircle[gold,fill=gold]{2pt} & 0.244~\tikzcircle[gold,fill=gold]{2pt} & 31.72~\tikzcircle[gold,fill=gold]{2pt} & 0.917~\tikzcircle[gold,fill=gold]{2pt} & 0.179~\tikzcircle[gold,fill=gold]{2pt} \\
  & Ours (full ray) & 24.30~\tikzcircle[silver,fill=silver]{2pt} & 0.703~\tikzcircle[silver,fill=silver]{2pt} & 0.316~\tikzcircle[silver,fill=silver]{2pt} & 29.04 & 0.900~\tikzcircle[silver,fill=silver]{2pt} & 0.239~\tikzcircle[silver,fill=silver]{2pt} \\
  \midrule
  \midrule
  \multirow{4}{*}{\rotatebox[origin=c]{90}{\emph{interactive}}}
  & I-NGP~\cite{mueller2022ingp} & 23.90~\tikzcircle[bronze,fill=bronze]{2pt} & 0.648~\tikzcircle[bronze,fill=bronze]{2pt} & 0.369 & 29.47~\tikzcircle[silver,fill=silver]{2pt}  & 0.877~\tikzcircle[bronze,fill=bronze]{2pt} & 0.273~\tikzcircle[bronze,fill=bronze]{2pt}   \\
  & MobileNeRF~\cite{chen2022mobilenerf}  & 22.90 & 0.524 & 0.463 & 25.74  & 0.757 & 0.453  \\
  & BakedSDF~\cite{yariv2023bakedsdf} & 23.40 & 0.577 & 0.351~\tikzcircle[bronze,fill=bronze]{2pt} & 27.20 & 0.845 & 0.300 \\
  & Ours & 23.17 & 0.606 & 0.389 & 29.19~\tikzcircle[bronze,fill=bronze]{2pt}	& 0.872 & 0.285 \\
  \bottomrule
\end{tabular}
}
\end{table}

\begin{table}[t!]
\centering
\small
\caption{Ablating our method on the \ShellyDataset{} data set. SV Kernel denotes the spatially varying kernel as introduced in Section~\ref{sec:extending_neus}. \textit{Band, fixed} denotes the shell is not adaptive but extracted for a given SDF threshold.}
\label{tab:shelly_ablation}
\vspace{-1em}
\begin{adjustbox}{width=.45\textwidth,center}
\begin{tabular}{l|cccc}
\toprule
Model                            & \multicolumn{1}{c}{PSNR}~$\uparrow$ & LPIPS~$\downarrow$ &SSIM~$\uparrow$ & Sample~$\downarrow$ \\ 
\midrule
Ours (full ray, w/o SV   kernel)    & 32.99    & 0.115   & 0.921      & 384     \\
Ours (full ray)                     & 34.26    & 0.104   & 0.932      & 384     \\
\midrule
Ours (band, fixed $\pm$0.05)    & 33.83     & 0.110     & 0.928      & 4.51      \\
Ours (band, fixed $\pm$0.02)    & 31.14     & 0.136     & 0.913      & 2.29      \\
Ours (keep regularization)      & 34.22     & 0.085     & 0.948      & 1.74      \\
Ours                            & 36.02     & 0.079     & 0.954      & 1.74      \\
\bottomrule
\end{tabular}
\end{adjustbox}
\end{table}
\begin{figure}
\begin{minipage}{0.24\textwidth}
  \centering
  \includegraphics[width=\linewidth]{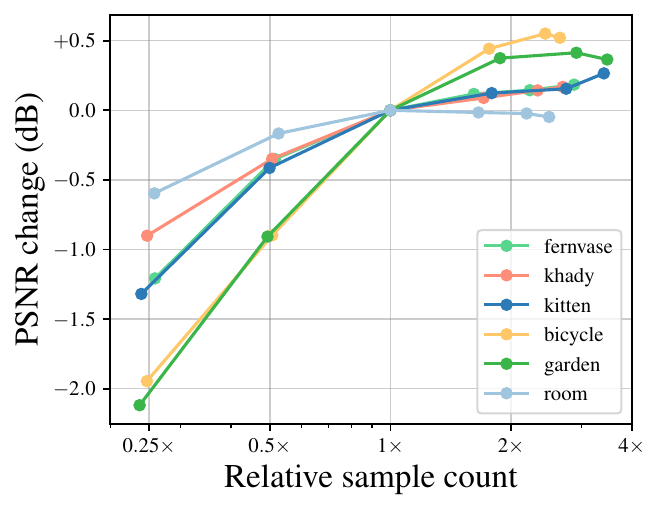}
  \label{fig:ablation_plot_psnr}
\end{minipage}%
\begin{minipage}{0.24\textwidth}
  \centering
  \includegraphics[width=\linewidth]{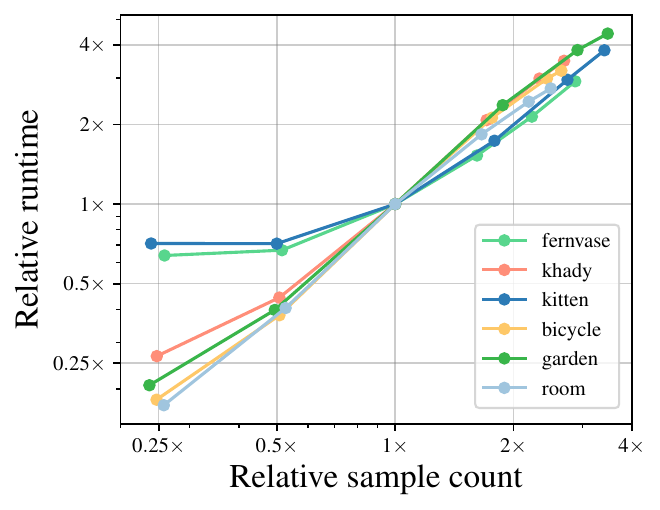}
  \label{fig:ablation_plot_runtime}
\end{minipage}
\vspace*{-1.5em}
\caption{
\revAdded{Ablating the effect of sample count on image quality and runtime performance.
We vary the sample count, and plot the PSNR change (left) and relative runtime performance (right) compared to the default hyperparameters denoted as ``$1\times$ sample count''.
We experiment with six scenes from the \ShellyDataset{} (fernvase, khady, kitten) and \MipNerfDataset{} (bicycle, garden, room) data sets.
}
}
\label{fig:ablation_plot}
\end{figure}

\subsection{Ablation Study}
We ablate our design choices on the \ShellyDataset{} data set in \tabref{shelly_ablation}. In line with our motivation in \secref{extending_neus}, the spatially-varying kernel size provides the required flexibility to adapt to the local complexity of the scene which results in improvement across all metrics. Using a fixed SDF threshold to extract the band requires seeking a compromise between an adaptive shell that is too narrow to represent fuzzy surfaces (threshold 0.02) or an increased sample count (threshold 0.05). Instead, our formulation can automatically adapt to the local complexity of the scene leading to higher quality metrics and lower sample count.
As described in \secref{training}, we disable the regularization terms after shell extraction to devote more capacity to fit the visual appearance. Comparing \emph{Ours (keep regularization)} with \emph{Ours}, this leads to improved visual fidelity.

In \figref{ablation_plot}, we ablate our method and study how image quality and runtime change with different sample counts.
We vary the sample step size $\IntervalSampleSpacing$ in narrow-band rendering (\secref{narrow_band_rendering}) to produce varying sample counts, and keep other hyperparameters unchanged.
The PSNR is sensitive to sample counts when the samples are insufficient ($0.25\times$-$1\times$), and the image quality starts to saturate as the sample counts go higher ($1\times$-$4\times$).
In most scenes, the runtime performance is approximately linear w.r.t. the sample count.
For simpler scenes such as \textsc{Kitten} and \textsc{Fernvase}, smaller sample counts ($0.25\times$-$1\times$) do not further reduce the runtime due to a mixture of fixed overheads (e.g. Python interpreter and Dr.Jit tracing) and under-utilization of the GPU.

\section{Applications}
\label{sec:applications}

Our method directly constructs an explicit outer shell mesh $\OuterMesh$ which by definition contains all regions of space that contribute to the rendered appearance. This property has great utility for use in downstream applications.

So far our scenes have represented entirely static content, yet, the world is full of motion.
Cage-based deformation methods have shown promise for enabling dynamic, non-rigid motion in NeRF and other volumetric representations~\cite{10.1145/1276377.1276466,lee2018skinned,yuan2022nerf,xu2022deforming,garbin2022voltemorph}.
The basic idea is to construct a coarse tetrahedral cage around a neural volume, deform the cage, and use it to render the deformed appearance of the underlying volume.
Our approach perfectly supports such techniques, as the outer shell mesh $\OuterMesh$ guides the construction of a cage which will surely contain the content.

We first dilate and tetrahedralize the outer mesh $\OuterMesh$ with FastTetWild~\cite{hu2020fast} to produce a tetrahedral mesh that encapsulates the scene.
This mesh acts as a proxy for performing physics simulations, animations, editing, and other operations. 
To render our representation after deforming the tetrahedral cage, any deformation is transferred to $\OuterMesh$ and $\InnerMesh$ via barycentric interpolation, using precomputed barycentric coordinates generated as a preprocess.
Ray directions are likewise transformed via finite differences.
After the transformation, we proceed with rendering as usual in the reference space of our representation, as described in \secref{narrow_band_rendering}.
Note that even in the presence of deformations, the rendering process still benefits from our efficient adaptive shell representation, and is able to efficiently sample the underlying neural volume.

We show two examples of applying physical simulation and animation to the reconstructed objects in \figref{animation}; please see the supplemental video for dynamic motion.
In the animation example, we manually drive the motion of plants in a vase according to an analytical wind-like spatial function.
Other animation schemes, such as blend shapes or character rigs could potentially be substituted to drive the motion.
In the physical simulation example, we simulate the reconstructed asset via finite-element elastic simulation on the cage tetrahedra including collision penalties~\cite{jatavallabhula2021gradsim}.

\begin{figure}
    \centering
    \includegraphics{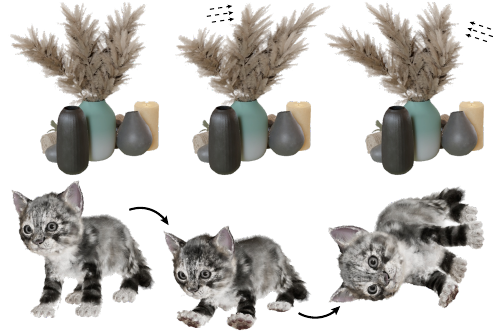}
    \caption{Our representation is well-suited for animation (\emph{top}) and physical simulation (bottom).
    The visual quality is preserved under deformation: the original shape is shown in leftmost column, with deformations in the middle and rightmost column.
    For details, please zoom into the fuzzy regions (e.g. fur, leaves), and refer to the supplemental video.}
    \label{fig:animation}
\end{figure}

\section{Discussion}

Recent work has developed schemes to accelerate and improve the quality of NeRF-like scene representations.
\secref{experiments} provides comparisons to selected, particularly relevant methods. Note that due to the high research activity in the field, it is impossible to compare to all techniques and for many approaches implementations are not available.
Hence, we offer additional comments on some related work:

\begin{itemize}[leftmargin=*]
  \item MobileNeRF~\cite{chen2022mobilenerf}, BakedSDF~\cite{yariv2023bakedsdf}, NeRFMeshing~\cite{rakotosaona2023nerfmeshing}, and nerf2mesh ~\cite{tang2023delicate} post-process NeRF-like models and extract meshes to accelerate inference, similar to this work. However, these approaches constrain appearance to surfaces, sacrificing quality.
  Our method instead retains a full volumetric representation and nearly full-NeRF quality, at the cost of moderately more expensive inference (though still real-time on modern hardware).

  \item DuplexRF~\cite{wan2023Duplex} also extracts an explicit shell from the underlying neural field and uses it to accelerate rendering, although it does so with a very different neural representation, prioritizing performance. Their shell is directly extracted from two thresholds of the radiance field, which requires the careful selection of the thresholds and results in a noisy shell that is not adapted to the local complexity of the scene in contrast to our approach.

  \item VMesh~\cite{guo2023vmesh} builds upon the similar insight that different parts of the scene require different treatment. However, their formulation assumes an additional voxel-grid data structure to mark the volumetric areas that contribute to the final rendering. This approach suffers from poor complexity scaling as with the auxiliary acceleration data structure of \cite{mueller2022ingp}. Instead, our method uses an explicit, adaptive shell to delimit the areas that contribute to the rendering. Apart from lower complexity, our formulation seamlessly enables further applications as discussed in \secref{applications}.

\end{itemize}

\section{Conclusion and Future Work}
In this work we focus on efficiently rendering NeRFs.
Our first stage of training (Section~\ref{sec:extending_neus}) is largely similar to that of ~\cite{Li2023Neuralangelo}, and likely can be accelerated by algorithmic advancements and low-level tuning similar to our inference pipeline~\cite{wang2022neus2}.

Although our method offers large speedups for high-fidelity neural rendering and runs at real-time rates on modern hardware (\tabref{perf_comparison}, it is still significantly more expensive than approaches such as MeRF~\cite{reiser2023merf} that precompute the neural field outputs and bake them onto a discrete grid representation. Our formulation is complimentary to that of MeRF~\cite{reiser2023merf} and we hypothesize that combining both approaches will lead to further speedups, potentially reaching the performance---at high quality---of the methods that bake the volumetric representation to explicit meshes and can run in real-time even on commodity hardware (\eg{}~\cite{chen2022mobilenerf}).

Our method does not guarantee to capture thin structures---if the extracted adaptive shell omits a geometric region, it can never be recovered during fine-tuning and will always be absent from the reconstruction.
Artifacts of this form are visible in some \MipNerfDataset{} scenes.
Future work will explore an iterative procedure, in which we alternately tune our reconstruction and adapt the shell to ensure that no significant geometry is missed.
Other artifacts occasionally present in our reconstructions include spurious floating geometry and poorly-resolved backgrounds; both are common challenges in neural reconstructions and our approach may borrow solutions from other work across the field (\eg{}~\cite{Niemeyer2021Regnerf}).

More broadly, there is great potential in combining recent neural representations with high-performance techniques honed for real-time performance in computer graphics. Here, we have shown how ray tracing and adaptive shells can be used to greatly improve performance.

\begin{acks}
The authors are grateful to Alex Evans for the insightful brainstorming sessions and his explorations in the early stages of this project. We also appreciate the feedback received from Jacob Munkberg, Jon Hasselgren, Chen-Hsuan Lin, and Wenzheng Chen during the project. Finally, we would like to thank Lior Yariv for providing the results of BakedSDF.
\end{acks}

{\small
\bibliographystyle{ACM-Reference-Format}
\bibliography{egbib}


\begin{thebibliography}{71}


\ifx \showCODEN    \undefined \def \showCODEN     #1{\unskip}     \fi
\ifx \showDOI      \undefined \def \showDOI       #1{#1}\fi
\ifx \showISBNx    \undefined \def \showISBNx     #1{\unskip}     \fi
\ifx \showISBNxiii \undefined \def \showISBNxiii  #1{\unskip}     \fi
\ifx \showISSN     \undefined \def \showISSN      #1{\unskip}     \fi
\ifx \showLCCN     \undefined \def \showLCCN      #1{\unskip}     \fi
\ifx \shownote     \undefined \def \shownote      #1{#1}          \fi
\ifx \showarticletitle \undefined \def \showarticletitle #1{#1}   \fi
\ifx \showURL      \undefined \def \showURL       {\relax}        \fi
\providecommand\bibfield[2]{#2}
\providecommand\bibinfo[2]{#2}
\providecommand\natexlab[1]{#1}
\providecommand\showeprint[2][]{arXiv:#2}

\bibitem[Barron et~al\mbox{.}(2021)]%
        {barron2021mipnerf}
\bibfield{author}{\bibinfo{person}{Jonathan~T. Barron}, \bibinfo{person}{Ben
  Mildenhall}, \bibinfo{person}{Matthew Tancik}, \bibinfo{person}{Peter
  Hedman}, \bibinfo{person}{Ricardo Martin-Brualla}, {and}
  \bibinfo{person}{Pratul~P. Srinivasan}.} \bibinfo{year}{2021}\natexlab{}.
\newblock \showarticletitle{Mip-NeRF: A Multiscale Representation for
  Anti-Aliasing Neural Radiance Fields}.
\newblock \bibinfo{journal}{\emph{ICCV}} (\bibinfo{year}{2021}).
\newblock


\bibitem[Barron et~al\mbox{.}(2022)]%
        {barron2022mipnerf360}
\bibfield{author}{\bibinfo{person}{Jonathan~T. Barron}, \bibinfo{person}{Ben
  Mildenhall}, \bibinfo{person}{Dor Verbin}, \bibinfo{person}{Pratul~P.
  Srinivasan}, {and} \bibinfo{person}{Peter Hedman}.}
  \bibinfo{year}{2022}\natexlab{}.
\newblock \showarticletitle{Mip-NeRF 360: Unbounded Anti-Aliased Neural
  Radiance Fields}.
\newblock \bibinfo{journal}{\emph{CVPR}} (\bibinfo{year}{2022}).
\newblock


\bibitem[Bi et~al\mbox{.}(2021)]%
        {bi2021deep}
\bibfield{author}{\bibinfo{person}{Sai Bi}, \bibinfo{person}{Stephen Lombardi},
  \bibinfo{person}{Shunsuke Saito}, \bibinfo{person}{Tomas Simon},
  \bibinfo{person}{Shih-En Wei}, \bibinfo{person}{Kevyn Mcphail},
  \bibinfo{person}{Ravi Ramamoorthi}, \bibinfo{person}{Yaser Sheikh}, {and}
  \bibinfo{person}{Jason Saragih}.} \bibinfo{year}{2021}\natexlab{}.
\newblock \showarticletitle{Deep relightable appearance models for animatable
  faces}.
\newblock \bibinfo{journal}{\emph{ACM Transactions on Graphics (TOG)}}
  \bibinfo{volume}{40}, \bibinfo{number}{4} (\bibinfo{year}{2021}),
  \bibinfo{pages}{1--15}.
\newblock


\bibitem[Buehler et~al\mbox{.}(2001)]%
        {Buehler2001ulr}
\bibfield{author}{\bibinfo{person}{Chris Buehler}, \bibinfo{person}{Michael
  Bosse}, \bibinfo{person}{Leonard McMillan}, \bibinfo{person}{Steven Gortler},
  {and} \bibinfo{person}{Michael Cohen}.} \bibinfo{year}{2001}\natexlab{}.
\newblock \showarticletitle{Unstructured Lumigraph Rendering}. In
  \bibinfo{booktitle}{\emph{Proceedings of the 28th Annual Conference on
  Computer Graphics and Interactive Techniques}}
  \emph{(\bibinfo{series}{SIGGRAPH '01})}. \bibinfo{publisher}{Association for
  Computing Machinery}, \bibinfo{address}{New York, NY, USA},
  \bibinfo{pages}{425–432}.
\newblock
\showISBNx{158113374X}


\bibitem[Cao et~al\mbox{.}(2022)]%
        {cao2022real}
\bibfield{author}{\bibinfo{person}{Junli Cao}, \bibinfo{person}{Huan Wang},
  \bibinfo{person}{Pavlo Chemerys}, \bibinfo{person}{Vladislav Shakhrai},
  \bibinfo{person}{Ju Hu}, \bibinfo{person}{Yun Fu}, \bibinfo{person}{Denys
  Makoviichuk}, \bibinfo{person}{Sergey Tulyakov}, {and} \bibinfo{person}{Jian
  Ren}.} \bibinfo{year}{2022}\natexlab{}.
\newblock \showarticletitle{Real-Time Neural Light Field on Mobile Devices}.
\newblock \bibinfo{journal}{\emph{arXiv preprint arXiv:2212.08057}}
  (\bibinfo{year}{2022}).
\newblock


\bibitem[Chen et~al\mbox{.}(2022)]%
        {Chen2022tensorrf}
\bibfield{author}{\bibinfo{person}{Anpei Chen}, \bibinfo{person}{Zexiang Xu},
  \bibinfo{person}{Andreas Geiger}, \bibinfo{person}{Jingyi Yu}, {and}
  \bibinfo{person}{Hao Su}.} \bibinfo{year}{2022}\natexlab{}.
\newblock \showarticletitle{TensoRF: Tensorial Radiance Fields}. In
  \bibinfo{booktitle}{\emph{European Conference on Computer Vision (ECCV)}}.
\newblock


\bibitem[Chen et~al\mbox{.}(2023)]%
        {chen2022mobilenerf}
\bibfield{author}{\bibinfo{person}{Zhiqin Chen}, \bibinfo{person}{Thomas
  Funkhouser}, \bibinfo{person}{Peter Hedman}, {and} \bibinfo{person}{Andrea
  Tagliasacchi}.} \bibinfo{year}{2023}\natexlab{}.
\newblock \showarticletitle{MobileNeRF: Exploiting the Polygon Rasterization
  Pipeline for Efficient Neural Field Rendering on Mobile Architectures}. In
  \bibinfo{booktitle}{\emph{The Conference on Computer Vision and Pattern
  Recognition (CVPR)}}.
\newblock


\bibitem[Davis et~al\mbox{.}(2012)]%
        {davis2012ulf}
\bibfield{author}{\bibinfo{person}{Abe Davis}, \bibinfo{person}{Marc Levoy},
  {and} \bibinfo{person}{Fredo Durand}.} \bibinfo{year}{2012}\natexlab{}.
\newblock \showarticletitle{Unstructured Light Fields}.
\newblock \bibinfo{journal}{\emph{Comput. Graph. Forum}} \bibinfo{volume}{31},
  \bibinfo{number}{2pt1} (\bibinfo{year}{2012}), \bibinfo{pages}{305–314}.
\newblock


\bibitem[Debevec et~al\mbox{.}(1996)]%
        {debevec1996hybrid}
\bibfield{author}{\bibinfo{person}{Paul~E. Debevec},
  \bibinfo{person}{Camillo~J. Taylor}, {and} \bibinfo{person}{Jitendra Malik}.}
  \bibinfo{year}{1996}\natexlab{}.
\newblock \showarticletitle{Modeling and Rendering Architecture from
  Photographs: A Hybrid Geometry- and Image-Based Approach}. In
  \bibinfo{booktitle}{\emph{Proceedings of the 23rd Annual Conference on
  Computer Graphics and Interactive Techniques}}
  \emph{(\bibinfo{series}{SIGGRAPH '96})}. \bibinfo{publisher}{Association for
  Computing Machinery}, \bibinfo{pages}{11–20}.
\newblock
\showISBNx{0897917464}


\bibitem[Deng et~al\mbox{.}(2022)]%
        {kangle2021dsnerf}
\bibfield{author}{\bibinfo{person}{Kangle Deng}, \bibinfo{person}{Andrew Liu},
  \bibinfo{person}{Jun-Yan Zhu}, {and} \bibinfo{person}{Deva Ramanan}.}
  \bibinfo{year}{2022}\natexlab{}.
\newblock \showarticletitle{Depth-supervised {NeRF}: Fewer Views and Faster
  Training for Free}. In \bibinfo{booktitle}{\emph{Proceedings of the IEEE/CVF
  Conference on Computer Vision and Pattern Recognition (CVPR)}}.
\newblock


\bibitem[Garbin et~al\mbox{.}(2022)]%
        {garbin2022voltemorph}
\bibfield{author}{\bibinfo{person}{Stephan~J Garbin}, \bibinfo{person}{Marek
  Kowalski}, \bibinfo{person}{Virginia Estellers}, \bibinfo{person}{Stanislaw
  Szymanowicz}, \bibinfo{person}{Shideh Rezaeifar}, \bibinfo{person}{Jingjing
  Shen}, \bibinfo{person}{Matthew Johnson}, {and} \bibinfo{person}{Julien
  Valentin}.} \bibinfo{year}{2022}\natexlab{}.
\newblock \showarticletitle{VolTeMorph: Realtime, Controllable and
  Generalisable Animation of Volumetric Representations}.
\newblock \bibinfo{journal}{\emph{arXiv preprint arXiv:2208.00949}}
  (\bibinfo{year}{2022}).
\newblock


\bibitem[Gortler et~al\mbox{.}(1996)]%
        {Gortler1996lumigraph}
\bibfield{author}{\bibinfo{person}{Steven~J. Gortler}, \bibinfo{person}{Radek
  Grzeszczuk}, \bibinfo{person}{Richard Szeliski}, {and}
  \bibinfo{person}{Michael~F. Cohen}.} \bibinfo{year}{1996}\natexlab{}.
\newblock \showarticletitle{The Lumigraph}. In
  \bibinfo{booktitle}{\emph{Proceedings of the 23rd Annual Conference on
  Computer Graphics and Interactive Techniques}}
  \emph{(\bibinfo{series}{SIGGRAPH '96})}. \bibinfo{publisher}{Association for
  Computing Machinery}, \bibinfo{pages}{43–54}.
\newblock
\showISBNx{0897917464}


\bibitem[Guo et~al\mbox{.}(2023)]%
        {guo2023vmesh}
\bibfield{author}{\bibinfo{person}{Yuan-Chen Guo}, \bibinfo{person}{Yan-Pei
  Cao}, \bibinfo{person}{Chen Wang}, \bibinfo{person}{Yu He},
  \bibinfo{person}{Ying Shan}, \bibinfo{person}{Xiaohu Qie}, {and}
  \bibinfo{person}{Song-Hai Zhang}.} \bibinfo{year}{2023}\natexlab{}.
\newblock \showarticletitle{VMesh: Hybrid Volume-Mesh Representation for
  Efficient View Synthesis}.
\newblock \bibinfo{journal}{\emph{arXiv preprint arXiv:2303.16184}}
  (\bibinfo{year}{2023}).
\newblock


\bibitem[Hedman et~al\mbox{.}(2021)]%
        {hedman2021snerg}
\bibfield{author}{\bibinfo{person}{Peter Hedman}, \bibinfo{person}{Pratul~P.
  Srinivasan}, \bibinfo{person}{Ben Mildenhall}, \bibinfo{person}{Jonathan~T.
  Barron}, {and} \bibinfo{person}{Paul Debevec}.}
  \bibinfo{year}{2021}\natexlab{}.
\newblock \showarticletitle{Baking Neural Radiance Fields for Real-Time View
  Synthesis}.
\newblock \bibinfo{journal}{\emph{ICCV}} (\bibinfo{year}{2021}).
\newblock


\bibitem[Hu et~al\mbox{.}(2022)]%
        {Hu2022EfficientNeRF}
\bibfield{author}{\bibinfo{person}{Tao Hu}, \bibinfo{person}{Shu Liu},
  \bibinfo{person}{Yilun Chen}, \bibinfo{person}{Tiancheng Shen}, {and}
  \bibinfo{person}{Jiaya Jia}.} \bibinfo{year}{2022}\natexlab{}.
\newblock \showarticletitle{EfficientNeRF Efficient Neural Radiance Fields}. In
  \bibinfo{booktitle}{\emph{Proceedings of the IEEE/CVF Conference on Computer
  Vision and Pattern Recognition (CVPR)}}. \bibinfo{pages}{12902--12911}.
\newblock


\bibitem[Hu et~al\mbox{.}(2020)]%
        {hu2020fast}
\bibfield{author}{\bibinfo{person}{Yixin Hu}, \bibinfo{person}{Teseo
  Schneider}, \bibinfo{person}{Bolun Wang}, \bibinfo{person}{Denis Zorin},
  {and} \bibinfo{person}{Daniele Panozzo}.} \bibinfo{year}{2020}\natexlab{}.
\newblock \showarticletitle{Fast tetrahedral meshing in the wild}.
\newblock \bibinfo{journal}{\emph{ACM Transactions on Graphics (TOG)}}
  \bibinfo{volume}{39}, \bibinfo{number}{4} (\bibinfo{year}{2020}),
  \bibinfo{pages}{117--1}.
\newblock


\bibitem[Jakob et~al\mbox{.}(2022)]%
        {Jakob2022DrJit}
\bibfield{author}{\bibinfo{person}{Wenzel Jakob},
  \bibinfo{person}{S{\'e}bastien Speierer}, \bibinfo{person}{Nicolas Roussel},
  {and} \bibinfo{person}{Delio Vicini}.} \bibinfo{year}{2022}\natexlab{}.
\newblock \showarticletitle{DR. JIT: a just-in-time compiler for differentiable
  rendering}.
\newblock \bibinfo{journal}{\emph{ACM Transactions on Graphics (TOG)}}
  \bibinfo{volume}{41}, \bibinfo{number}{4} (\bibinfo{year}{2022}),
  \bibinfo{pages}{1--19}.
\newblock


\bibitem[Jatavallabhula et~al\mbox{.}(2021)]%
        {jatavallabhula2021gradsim}
\bibfield{author}{\bibinfo{person}{Krishna~Murthy Jatavallabhula},
  \bibinfo{person}{Miles Macklin}, \bibinfo{person}{Florian Golemo},
  \bibinfo{person}{Vikram Voleti}, \bibinfo{person}{Linda Petrini},
  \bibinfo{person}{Martin Weiss}, \bibinfo{person}{Breandan Considine},
  \bibinfo{person}{J{\'e}r{\^o}me Parent-L{\'e}vesque}, \bibinfo{person}{Kevin
  Xie}, \bibinfo{person}{Kenny Erleben}, {et~al\mbox{.}}}
  \bibinfo{year}{2021}\natexlab{}.
\newblock \showarticletitle{gradsim: Differentiable simulation for system
  identification and visuomotor control}.
\newblock \bibinfo{journal}{\emph{arXiv preprint arXiv:2104.02646}}
  (\bibinfo{year}{2021}).
\newblock


\bibitem[Jensen et~al\mbox{.}(2014)]%
        {jensen2014large}
\bibfield{author}{\bibinfo{person}{Rasmus Jensen}, \bibinfo{person}{Anders
  Dahl}, \bibinfo{person}{George Vogiatzis}, \bibinfo{person}{Engil Tola},
  {and} \bibinfo{person}{Henrik Aan{\ae}s}.} \bibinfo{year}{2014}\natexlab{}.
\newblock \showarticletitle{Large scale multi-view stereopsis evaluation}. In
  \bibinfo{booktitle}{\emph{2014 IEEE Conference on Computer Vision and Pattern
  Recognition}}. IEEE, \bibinfo{pages}{406--413}.
\newblock


\bibitem[Joshi et~al\mbox{.}(2007)]%
        {10.1145/1276377.1276466}
\bibfield{author}{\bibinfo{person}{Pushkar Joshi}, \bibinfo{person}{Mark
  Meyer}, \bibinfo{person}{Tony DeRose}, \bibinfo{person}{Brian Green}, {and}
  \bibinfo{person}{Tom Sanocki}.} \bibinfo{year}{2007}\natexlab{}.
\newblock \showarticletitle{Harmonic Coordinates for Character Articulation}.
\newblock \bibinfo{journal}{\emph{ACM Trans. Graph.}} \bibinfo{volume}{26},
  \bibinfo{number}{3} (\bibinfo{date}{jul} \bibinfo{year}{2007}),
  \bibinfo{pages}{71–es}.
\newblock
\showISSN{0730-0301}
\urldef\tempurl%
\url{https://doi.org/10.1145/1276377.1276466}
\showDOI{\tempurl}


\bibitem[Karnewar et~al\mbox{.}(2022)]%
        {karnewar2022relufields}
\bibfield{author}{\bibinfo{person}{Animesh Karnewar}, \bibinfo{person}{Tobias
  Ritschel}, \bibinfo{person}{Oliver Wang}, {and} \bibinfo{person}{Niloy
  Mitra}.} \bibinfo{year}{2022}\natexlab{}.
\newblock \showarticletitle{ReLU Fields: The Little Non-Linearity That Could}.
  In \bibinfo{booktitle}{\emph{ACM SIGGRAPH 2022 Conference Proceedings}}
  \emph{(\bibinfo{series}{SIGGRAPH '22})}. \bibinfo{publisher}{Association for
  Computing Machinery}, \bibinfo{address}{New York, NY, USA}, Article
  \bibinfo{articleno}{27}, \bibinfo{numpages}{9}~pages.
\newblock
\showISBNx{9781450393379}
\urldef\tempurl%
\url{https://doi.org/10.1145/3528233.3530707}
\showDOI{\tempurl}


\bibitem[Kazhdan et~al\mbox{.}(2006)]%
        {kazdhan2006psr}
\bibfield{author}{\bibinfo{person}{Michael~M. Kazhdan},
  \bibinfo{person}{Matthew Bolitho}, {and} \bibinfo{person}{Hugues Hoppe}.}
  \bibinfo{year}{2006}\natexlab{}.
\newblock \showarticletitle{Poisson Surface Reconstruction}. In
  \bibinfo{booktitle}{\emph{Proceedings of the Fourth Eurographics Symposium on
  Geometry Processing}} \emph{(\bibinfo{series}{SGP '06},
  Vol.~\bibinfo{volume}{256})}. \bibinfo{publisher}{Eurographics Association},
  \bibinfo{pages}{61--70}.
\newblock


\bibitem[Kazhdan and Hoppe(2013)]%
        {kazdhan2013spsr}
\bibfield{author}{\bibinfo{person}{Michael~M. Kazhdan} {and}
  \bibinfo{person}{Hugues Hoppe}.} \bibinfo{year}{2013}\natexlab{}.
\newblock \showarticletitle{Screened poisson surface reconstruction.}
\newblock \bibinfo{journal}{\emph{ACM Trans. Graph.}} \bibinfo{volume}{32},
  \bibinfo{number}{3} (\bibinfo{year}{2013}), \bibinfo{pages}{29:1--29:13}.
\newblock


\bibitem[Kerbl et~al\mbox{.}(2023)]%
        {kerbl3Dgaussians}
\bibfield{author}{\bibinfo{person}{Bernhard Kerbl}, \bibinfo{person}{Georgios
  Kopanas}, \bibinfo{person}{Thomas Leimk{\"u}hler}, {and}
  \bibinfo{person}{George Drettakis}.} \bibinfo{year}{2023}\natexlab{}.
\newblock \showarticletitle{3D Gaussian Splatting for Real-Time Radiance Field
  Rendering}.
\newblock \bibinfo{journal}{\emph{ACM Transactions on Graphics}}
  \bibinfo{volume}{42}, \bibinfo{number}{4} (\bibinfo{date}{July}
  \bibinfo{year}{2023}).
\newblock
\urldef\tempurl%
\url{https://repo-sam.inria.fr/fungraph/3d-gaussian-splatting/}
\showURL{%
\tempurl}


\bibitem[Kopanas et~al\mbox{.}(2021)]%
        {Kopanas2021pointbasednr}
\bibfield{author}{\bibinfo{person}{Georgios Kopanas}, \bibinfo{person}{Julien
  Philip}, \bibinfo{person}{Thomas Leimkühler}, {and} \bibinfo{person}{George
  Drettakis}.} \bibinfo{year}{2021}\natexlab{}.
\newblock \showarticletitle{Point-Based Neural Rendering with Per-View
  Optimization}.
\newblock \bibinfo{journal}{\emph{Computer Graphics Forum (Proceedings of the
  Eurographics Symposium on Rendering)}} \bibinfo{volume}{40},
  \bibinfo{number}{4} (\bibinfo{date}{June} \bibinfo{year}{2021}).
\newblock
\urldef\tempurl%
\url{http://www-sop.inria.fr/reves/Basilic/2021/KPLD21}
\showURL{%
\tempurl}


\bibitem[Kurz et~al\mbox{.}(2022)]%
        {kurz2022adanerf}
\bibfield{author}{\bibinfo{person}{Andreas Kurz}, \bibinfo{person}{Thomas
  Neff}, \bibinfo{person}{Zhaoyang Lv}, \bibinfo{person}{Michael
  Zollh\"{o}fer}, {and} \bibinfo{person}{Markus Steinberger}.}
  \bibinfo{year}{2022}\natexlab{}.
\newblock \showarticletitle{AdaNeRF: Adaptive Sampling for Real-time Rendering
  of Neural Radiance Fields}. In \bibinfo{booktitle}{\emph{European Conference
  on Computer Vision (ECCV)}}.
\newblock


\bibitem[Lassner and Zollh{\"o}fer(2021)]%
        {Lassner2021PulsarES}
\bibfield{author}{\bibinfo{person}{Christoph Lassner} {and}
  \bibinfo{person}{Michael Zollh{\"o}fer}.} \bibinfo{year}{2021}\natexlab{}.
\newblock \showarticletitle{Pulsar: Efficient Sphere-based Neural Rendering}.
\newblock \bibinfo{journal}{\emph{2021 IEEE/CVF Conference on Computer Vision
  and Pattern Recognition (CVPR)}} (\bibinfo{year}{2021}),
  \bibinfo{pages}{1440--1449}.
\newblock


\bibitem[Lee et~al\mbox{.}(2018)]%
        {lee2018skinned}
\bibfield{author}{\bibinfo{person}{Minjae Lee}, \bibinfo{person}{David Hyde},
  \bibinfo{person}{Michael Bao}, {and} \bibinfo{person}{Ronald Fedkiw}.}
  \bibinfo{year}{2018}\natexlab{}.
\newblock \showarticletitle{A skinned tetrahedral mesh for hair animation and
  hair-water interaction}.
\newblock \bibinfo{journal}{\emph{IEEE transactions on visualization and
  computer graphics}} \bibinfo{volume}{25}, \bibinfo{number}{3}
  (\bibinfo{year}{2018}), \bibinfo{pages}{1449--1459}.
\newblock


\bibitem[Levoy and Hanrahan(1996)]%
        {Levoy1996lightfields}
\bibfield{author}{\bibinfo{person}{Marc Levoy} {and} \bibinfo{person}{Pat
  Hanrahan}.} \bibinfo{year}{1996}\natexlab{}.
\newblock \showarticletitle{Light Field Rendering}. In
  \bibinfo{booktitle}{\emph{Proceedings of the 23rd Annual Conference on
  Computer Graphics and Interactive Techniques}}
  \emph{(\bibinfo{series}{SIGGRAPH '96})}. \bibinfo{publisher}{Association for
  Computing Machinery}, \bibinfo{pages}{31–42}.
\newblock
\showISBNx{0897917464}


\bibitem[Li et~al\mbox{.}(2023)]%
        {Li2023Neuralangelo}
\bibfield{author}{\bibinfo{person}{Max~Zhaoshuo Li}, \bibinfo{person}{Thomas
  M\"uller}, \bibinfo{person}{Alex Evans}, \bibinfo{person}{Russell~H. Taylor},
  \bibinfo{person}{Mathias Unberath}, \bibinfo{person}{Ming-Yu Liu}, {and}
  \bibinfo{person}{Chen-Hsuan Lin}.} \bibinfo{year}{2023}\natexlab{}.
\newblock \showarticletitle{Neuralangelo: High-Fidelity Neural Surface
  Reconstruction}. In \bibinfo{booktitle}{\emph{Conference on Computer Vision
  and Pattern Recognition (CVPR)}}.
\newblock


\bibitem[Lin et~al\mbox{.}(2022)]%
        {lin2022efficientnerf}
\bibfield{author}{\bibinfo{person}{Haotong Lin}, \bibinfo{person}{Sida Peng},
  \bibinfo{person}{Zhen Xu}, \bibinfo{person}{Yunzhi Yan},
  \bibinfo{person}{Qing Shuai}, \bibinfo{person}{Hujun Bao}, {and}
  \bibinfo{person}{Xiaowei Zhou}.} \bibinfo{year}{2022}\natexlab{}.
\newblock \showarticletitle{Efficient Neural Radiance Fields with Learned
  Depth-Guided Sampling}. In \bibinfo{booktitle}{\emph{SIGGRAPH Asia Conference
  Proceedings}}.
\newblock


\bibitem[Liu et~al\mbox{.}(2020)]%
        {liu2020nsvf}
\bibfield{author}{\bibinfo{person}{Lingjie Liu}, \bibinfo{person}{Jiatao Gu},
  \bibinfo{person}{Kyaw~Zaw Lin}, \bibinfo{person}{Tat-Seng Chua}, {and}
  \bibinfo{person}{Christian Theobalt}.} \bibinfo{year}{2020}\natexlab{}.
\newblock \showarticletitle{Neural Sparse Voxel Fields}.
\newblock \bibinfo{journal}{\emph{NeurIPS}} (\bibinfo{year}{2020}).
\newblock


\bibitem[Martin-Brualla et~al\mbox{.}(2021)]%
        {martinbrualla2020nerfw}
\bibfield{author}{\bibinfo{person}{Ricardo Martin-Brualla},
  \bibinfo{person}{Noha Radwan}, \bibinfo{person}{Mehdi S.~M. Sajjadi},
  \bibinfo{person}{Jonathan~T. Barron}, \bibinfo{person}{Alexey Dosovitskiy},
  {and} \bibinfo{person}{Daniel Duckworth}.} \bibinfo{year}{2021}\natexlab{}.
\newblock \showarticletitle{{NeRF in the Wild: Neural Radiance Fields for
  Unconstrained Photo Collections}}. In \bibinfo{booktitle}{\emph{CVPR}}.
\newblock


\bibitem[Mildenhall et~al\mbox{.}(2020)]%
        {mildenhall2020nerf}
\bibfield{author}{\bibinfo{person}{Ben Mildenhall}, \bibinfo{person}{Pratul~P.
  Srinivasan}, \bibinfo{person}{Matthew Tancik}, \bibinfo{person}{Jonathan~T.
  Barron}, \bibinfo{person}{Ravi Ramamoorthi}, {and} \bibinfo{person}{Ren Ng}.}
  \bibinfo{year}{2020}\natexlab{}.
\newblock \showarticletitle{NeRF: Representing Scenes as Neural Radiance Fields
  for View Synthesis}. In \bibinfo{booktitle}{\emph{ECCV}}.
\newblock


\bibitem[M\"uller et~al\mbox{.}(2022)]%
        {mueller2022ingp}
\bibfield{author}{\bibinfo{person}{Thomas M\"uller}, \bibinfo{person}{Alex
  Evans}, \bibinfo{person}{Christoph Schied}, {and} \bibinfo{person}{Alexander
  Keller}.} \bibinfo{year}{2022}\natexlab{}.
\newblock \showarticletitle{Instant Neural Graphics Primitives with a
  Multiresolution Hash Encoding}.
\newblock \bibinfo{journal}{\emph{ACM Trans. Graph.}} \bibinfo{volume}{41},
  \bibinfo{number}{4}, Article \bibinfo{articleno}{102} (\bibinfo{date}{July}
  \bibinfo{year}{2022}), \bibinfo{numpages}{15}~pages.
\newblock
\urldef\tempurl%
\url{https://doi.org/10.1145/3528223.3530127}
\showDOI{\tempurl}


\bibitem[Munkberg et~al\mbox{.}(2022)]%
        {Munkberg2022nvdiffrec}
\bibfield{author}{\bibinfo{person}{Jacob Munkberg}, \bibinfo{person}{Jon
  Hasselgren}, \bibinfo{person}{Tianchang Shen}, \bibinfo{person}{Jun Gao},
  \bibinfo{person}{Wenzheng Chen}, \bibinfo{person}{Alex Evans},
  \bibinfo{person}{Thomas M\"uller}, {and} \bibinfo{person}{Sanja Fidler}.}
  \bibinfo{year}{2022}\natexlab{}.
\newblock \showarticletitle{{Extracting Triangular 3D Models, Materials, and
  Lighting From Images}}. In \bibinfo{booktitle}{\emph{Proceedings of the
  IEEE/CVF Conference on Computer Vision and Pattern Recognition (CVPR)}}.
  \bibinfo{pages}{8280--8290}.
\newblock


\bibitem[Neff et~al\mbox{.}(2021)]%
        {neff2021donerf}
\bibfield{author}{\bibinfo{person}{Thomas Neff}, \bibinfo{person}{Pascal
  Stadlbauer}, \bibinfo{person}{Mathias Parger}, \bibinfo{person}{Andreas
  Kurz}, \bibinfo{person}{Joerg~H. Mueller}, \bibinfo{person}{Chakravarty
  R.~Alla Chaitanya}, \bibinfo{person}{Anton~S. Kaplanyan}, {and}
  \bibinfo{person}{Markus Steinberger}.} \bibinfo{year}{2021}\natexlab{}.
\newblock \showarticletitle{{DONeRF: Towards Real-Time Rendering of Compact
  Neural Radiance Fields using Depth Oracle Networks}}.
\newblock \bibinfo{journal}{\emph{Computer Graphics Forum}}
  \bibinfo{volume}{40}, \bibinfo{number}{4} (\bibinfo{year}{2021}).
\newblock


\bibitem[Niemeyer et~al\mbox{.}(2022)]%
        {Niemeyer2021Regnerf}
\bibfield{author}{\bibinfo{person}{Michael Niemeyer},
  \bibinfo{person}{Jonathan~T. Barron}, \bibinfo{person}{Ben Mildenhall},
  \bibinfo{person}{Mehdi S.~M. Sajjadi}, \bibinfo{person}{Andreas Geiger},
  {and} \bibinfo{person}{Noha Radwan}.} \bibinfo{year}{2022}\natexlab{}.
\newblock \showarticletitle{RegNeRF: Regularizing Neural Radiance Fields for
  View Synthesis from Sparse Inputs}. In \bibinfo{booktitle}{\emph{Proc. IEEE
  Conf. on Computer Vision and Pattern Recognition (CVPR)}}.
\newblock


\bibitem[Oechsle et~al\mbox{.}(2021)]%
        {Oechsle2021unisurf}
\bibfield{author}{\bibinfo{person}{Michael Oechsle}, \bibinfo{person}{Songyou
  Peng}, {and} \bibinfo{person}{Andreas Geiger}.}
  \bibinfo{year}{2021}\natexlab{}.
\newblock \showarticletitle{UNISURF: Unifying Neural Implicit Surfaces and
  Radiance Fields for Multi-View Reconstruction}. In
  \bibinfo{booktitle}{\emph{International Conference on Computer Vision
  (ICCV)}}.
\newblock


\bibitem[Parker et~al\mbox{.}(2010)]%
        {parker2010optix}
\bibfield{author}{\bibinfo{person}{Steven~G. Parker}, \bibinfo{person}{James
  Bigler}, \bibinfo{person}{Andreas Dietrich}, \bibinfo{person}{Heiko
  Friedrich}, \bibinfo{person}{Jared Hoberock}, \bibinfo{person}{David Luebke},
  \bibinfo{person}{David McAllister}, \bibinfo{person}{Morgan McGuire},
  \bibinfo{person}{Keith Morley}, \bibinfo{person}{Austin Robison}, {and}
  \bibinfo{person}{Martin Stich}.} \bibinfo{year}{2010}\natexlab{}.
\newblock \showarticletitle{OptiX: A General Purpose Ray Tracing Engine}.
\newblock \bibinfo{journal}{\emph{ACM Trans. Graph.}} \bibinfo{volume}{29},
  \bibinfo{number}{4}, Article \bibinfo{articleno}{66} (\bibinfo{date}{jul}
  \bibinfo{year}{2010}), \bibinfo{numpages}{13}~pages.
\newblock
\showISSN{0730-0301}
\urldef\tempurl%
\url{https://doi.org/10.1145/1778765.1778803}
\showDOI{\tempurl}


\bibitem[Paszke et~al\mbox{.}(2017)]%
        {paszke2017pytorch}
\bibfield{author}{\bibinfo{person}{Adam Paszke}, \bibinfo{person}{Sam Gross},
  \bibinfo{person}{Soumith Chintala}, \bibinfo{person}{Gregory Chanan},
  \bibinfo{person}{Edward Yang}, \bibinfo{person}{Zachary DeVito},
  \bibinfo{person}{Zeming Lin}, \bibinfo{person}{Alban Desmaison},
  \bibinfo{person}{Luca Antiga}, {and} \bibinfo{person}{Adam Lerer}.}
  \bibinfo{year}{2017}\natexlab{}.
\newblock \showarticletitle{Automatic differentiation in PyTorch}.
\newblock \bibinfo{journal}{\emph{NeurIPS Workshop on Autodiff}}
  (\bibinfo{year}{2017}).
\newblock


\bibitem[Rakotosaona et~al\mbox{.}(2023)]%
        {rakotosaona2023nerfmeshing}
\bibfield{author}{\bibinfo{person}{Marie-Julie Rakotosaona},
  \bibinfo{person}{Fabian Manhardt}, \bibinfo{person}{Diego~Martin Arroyo},
  \bibinfo{person}{Michael Niemeyer}, \bibinfo{person}{Abhijit Kundu}, {and}
  \bibinfo{person}{Federico Tombari}.} \bibinfo{year}{2023}\natexlab{}.
\newblock \showarticletitle{NeRFMeshing: Distilling Neural Radiance Fields into
  Geometrically-Accurate 3D Meshes}.
\newblock \bibinfo{journal}{\emph{arXiv preprint arXiv:2303.09431}}
  (\bibinfo{year}{2023}).
\newblock


\bibitem[Rebain et~al\mbox{.}(2020)]%
        {Rebain2021DeRF}
\bibfield{author}{\bibinfo{person}{Daniel Rebain}, \bibinfo{person}{Wei Jiang},
  \bibinfo{person}{Soroosh Yazdani}, \bibinfo{person}{Ke Li},
  \bibinfo{person}{Kwang~Moo Yi}, {and} \bibinfo{person}{Andrea Tagliasacchi}.}
  \bibinfo{year}{2020}\natexlab{}.
\newblock \showarticletitle{DeRF: Decomposed Radiance Fields}.
\newblock \bibinfo{journal}{\emph{2021 IEEE/CVF Conference on Computer Vision
  and Pattern Recognition (CVPR)}} (\bibinfo{year}{2020}),
  \bibinfo{pages}{14148--14156}.
\newblock


\bibitem[Reiser et~al\mbox{.}(2021)]%
        {Reiser2021kilonerf}
\bibfield{author}{\bibinfo{person}{Christian Reiser}, \bibinfo{person}{Songyou
  Peng}, \bibinfo{person}{Yiyi Liao}, {and} \bibinfo{person}{Andreas Geiger}.}
  \bibinfo{year}{2021}\natexlab{}.
\newblock \showarticletitle{KiloNeRF: Speeding up Neural Radiance Fields with
  Thousands of Tiny MLPs}. In \bibinfo{booktitle}{\emph{International
  Conference on Computer Vision (ICCV)}}.
\newblock


\bibitem[Reiser et~al\mbox{.}(2023)]%
        {reiser2023merf}
\bibfield{author}{\bibinfo{person}{Christian Reiser}, \bibinfo{person}{Richard
  Szeliski}, \bibinfo{person}{Dor Verbin}, \bibinfo{person}{Pratul~P
  Srinivasan}, \bibinfo{person}{Ben Mildenhall}, \bibinfo{person}{Andreas
  Geiger}, \bibinfo{person}{Jonathan~T Barron}, {and} \bibinfo{person}{Peter
  Hedman}.} \bibinfo{year}{2023}\natexlab{}.
\newblock \showarticletitle{Merf: Memory-efficient radiance fields for
  real-time view synthesis in unbounded scenes}.
\newblock \bibinfo{journal}{\emph{arXiv preprint arXiv:2302.12249}}
  (\bibinfo{year}{2023}).
\newblock


\bibitem[Riegler and Koltun(2020)]%
        {Riegler2020FVS}
\bibfield{author}{\bibinfo{person}{Gernot Riegler} {and}
  \bibinfo{person}{Vladlen Koltun}.} \bibinfo{year}{2020}\natexlab{}.
\newblock \showarticletitle{Free View Synthesis}. In
  \bibinfo{booktitle}{\emph{European Conference on Computer Vision}}.
\newblock


\bibitem[Riegler and Koltun(2021)]%
        {Riegler2021SVS}
\bibfield{author}{\bibinfo{person}{Gernot Riegler} {and}
  \bibinfo{person}{Vladlen Koltun}.} \bibinfo{year}{2021}\natexlab{}.
\newblock \showarticletitle{Stable View Synthesis}. In
  \bibinfo{booktitle}{\emph{Proceedings of the IEEE Conference on Computer
  Vision and Pattern Recognition}}.
\newblock


\bibitem[Rosu and Behnke(2023)]%
        {rosu2023permutosdf}
\bibfield{author}{\bibinfo{person}{Radu~Alexandru Rosu} {and}
  \bibinfo{person}{Sven Behnke}.} \bibinfo{year}{2023}\natexlab{}.
\newblock \showarticletitle{PermutoSDF: Fast Multi-View Reconstruction with
  Implicit Surfaces using Permutohedral Lattices}. In
  \bibinfo{booktitle}{\emph{IEEE/CVF Conference on Computer Vision and Pattern
  Recognition (CVPR)}}, Vol.~\bibinfo{volume}{2}.
\newblock


\bibitem[R{\"u}ckert et~al\mbox{.}(2021)]%
        {ruckert2021adop}
\bibfield{author}{\bibinfo{person}{Darius R{\"u}ckert}, \bibinfo{person}{Linus
  Franke}, {and} \bibinfo{person}{Marc Stamminger}.}
  \bibinfo{year}{2021}\natexlab{}.
\newblock \showarticletitle{Adop: Approximate differentiable one-pixel point
  rendering}.
\newblock \bibinfo{journal}{\emph{arXiv preprint arXiv:2110.06635}}
  (\bibinfo{year}{2021}).
\newblock


\bibitem[{Sara Fridovich-Keil and Alex Yu} et~al\mbox{.}(2022)]%
        {yu_and_fridovichkeil2021plenoxels}
\bibfield{author}{\bibinfo{person}{{Sara Fridovich-Keil and Alex Yu}},
  \bibinfo{person}{Matthew Tancik}, \bibinfo{person}{Qinhong Chen},
  \bibinfo{person}{Benjamin Recht}, {and} \bibinfo{person}{Angjoo Kanazawa}.}
  \bibinfo{year}{2022}\natexlab{}.
\newblock \showarticletitle{Plenoxels: Radiance Fields without Neural
  Networks}. In \bibinfo{booktitle}{\emph{CVPR}}.
\newblock


\bibitem[Sch\"{o}nberger and Frahm(2016)]%
        {schoenberger2016sfm}
\bibfield{author}{\bibinfo{person}{Johannes~Lutz Sch\"{o}nberger} {and}
  \bibinfo{person}{Jan-Michael Frahm}.} \bibinfo{year}{2016}\natexlab{}.
\newblock \showarticletitle{Structure-from-Motion Revisited}. In
  \bibinfo{booktitle}{\emph{Conference on Computer Vision and Pattern
  Recognition (CVPR)}}.
\newblock


\bibitem[Sch\"{o}nberger et~al\mbox{.}(2016)]%
        {schoenberger2016mvs}
\bibfield{author}{\bibinfo{person}{Johannes~Lutz Sch\"{o}nberger},
  \bibinfo{person}{Enliang Zheng}, \bibinfo{person}{Marc Pollefeys}, {and}
  \bibinfo{person}{Jan-Michael Frahm}.} \bibinfo{year}{2016}\natexlab{}.
\newblock \showarticletitle{Pixelwise View Selection for Unstructured
  Multi-View Stereo}. In \bibinfo{booktitle}{\emph{European Conference on
  Computer Vision (ECCV)}}.
\newblock


\bibitem[Sun et~al\mbox{.}(2022)]%
        {SunSC22DVGO}
\bibfield{author}{\bibinfo{person}{Cheng Sun}, \bibinfo{person}{Min Sun}, {and}
  \bibinfo{person}{Hwann{-}Tzong Chen}.} \bibinfo{year}{2022}\natexlab{}.
\newblock \showarticletitle{Direct Voxel Grid Optimization: Super-fast
  Convergence for Radiance Fields Reconstruction}. In
  \bibinfo{booktitle}{\emph{CVPR}}.
\newblock


\bibitem[Tancik et~al\mbox{.}(2023)]%
        {nerfstudio}
\bibfield{author}{\bibinfo{person}{Matthew Tancik}, \bibinfo{person}{Ethan
  Weber}, \bibinfo{person}{Evonne Ng}, \bibinfo{person}{Ruilong Li},
  \bibinfo{person}{Brent Yi}, \bibinfo{person}{Justin Kerr},
  \bibinfo{person}{Terrance Wang}, \bibinfo{person}{Alexander Kristoffersen},
  \bibinfo{person}{Jake Austin}, \bibinfo{person}{Kamyar Salahi},
  \bibinfo{person}{Abhik Ahuja}, \bibinfo{person}{David McAllister}, {and}
  \bibinfo{person}{Angjoo Kanazawa}.} \bibinfo{year}{2023}\natexlab{}.
\newblock \showarticletitle{Nerfstudio: A Modular Framework for Neural Radiance
  Field Development}. In \bibinfo{booktitle}{\emph{ACM SIGGRAPH 2023 Conference
  Proceedings}} \emph{(\bibinfo{series}{SIGGRAPH '23})}.
\newblock


\bibitem[Tang et~al\mbox{.}(2023)]%
        {tang2023delicate}
\bibfield{author}{\bibinfo{person}{Jiaxiang Tang}, \bibinfo{person}{Hang Zhou},
  \bibinfo{person}{Xiaokang Chen}, \bibinfo{person}{Tianshu Hu},
  \bibinfo{person}{Errui Ding}, \bibinfo{person}{Jingdong Wang}, {and}
  \bibinfo{person}{Gang Zeng}.} \bibinfo{year}{2023}\natexlab{}.
\newblock \showarticletitle{Delicate textured mesh recovery from nerf via
  adaptive surface refinement}.
\newblock \bibinfo{journal}{\emph{arXiv preprint arXiv:2303.02091}}
  (\bibinfo{year}{2023}).
\newblock


\bibitem[Waechter et~al\mbox{.}(2014)]%
        {Fleet2014letcolor}
\bibfield{author}{\bibinfo{person}{Michael Waechter}, \bibinfo{person}{Nils
  Moehrle}, {and} \bibinfo{person}{Michael Goesele}.}
  \bibinfo{year}{2014}\natexlab{}.
\newblock \showarticletitle{Let There Be Color! Large-Scale Texturing of 3D
  Reconstructions}. In \bibinfo{booktitle}{\emph{Computer Vision -- ECCV 2014}}
  (Heidelberg) \emph{(\bibinfo{series}{Lecture Notes in Computer Science},
  Vol.~\bibinfo{volume}{8693})}, \bibfield{editor}{\bibinfo{person}{David
  Fleet}, \bibinfo{person}{Tomas Pajdla}, \bibinfo{person}{Bernt Schiele},
  {and} \bibinfo{person}{Tinne Tuytelaars}} (Eds.).
  \bibinfo{publisher}{Springer}, \bibinfo{pages}{836--850}.
\newblock
\showISBNx{978-3-319-10601-4}
\urldef\tempurl%
\url{https://doi.org/10.1007/978-3-319-10602-1_54}
\showDOI{\tempurl}


\bibitem[Wan et~al\mbox{.}(2023)]%
        {wan2023Duplex}
\bibfield{author}{\bibinfo{person}{Ziyu Wan}, \bibinfo{person}{Christian
  Richardt}, \bibinfo{person}{Alja{\v{z}} Bo{\v{z}}i{\v{c}}},
  \bibinfo{person}{Chao Li}, \bibinfo{person}{Vijay Rengarajan},
  \bibinfo{person}{Seonghyeon Nam}, \bibinfo{person}{Xiaoyu Xiang},
  \bibinfo{person}{Tuotuo Li}, \bibinfo{person}{Bo Zhu},
  \bibinfo{person}{Rakesh Ranjan}, {et~al\mbox{.}}}
  \bibinfo{year}{2023}\natexlab{}.
\newblock \showarticletitle{Learning Neural Duplex Radiance Fields for
  Real-Time View Synthesis}.
\newblock \bibinfo{journal}{\emph{arXiv preprint arXiv:2304.10537}}
  (\bibinfo{year}{2023}).
\newblock


\bibitem[Wang et~al\mbox{.}(2021)]%
        {wang2021neus}
\bibfield{author}{\bibinfo{person}{Peng Wang}, \bibinfo{person}{Lingjie Liu},
  \bibinfo{person}{Yuan Liu}, \bibinfo{person}{Christian Theobalt},
  \bibinfo{person}{Taku Komura}, {and} \bibinfo{person}{Wenping Wang}.}
  \bibinfo{year}{2021}\natexlab{}.
\newblock \showarticletitle{NeuS: Learning Neural Implicit Surfaces by Volume
  Rendering for Multi-view Reconstruction}.
\newblock \bibinfo{journal}{\emph{NeurIPS}} (\bibinfo{year}{2021}).
\newblock


\bibitem[Wang et~al\mbox{.}(2022a)]%
        {wang2022neus2}
\bibfield{author}{\bibinfo{person}{Yiming Wang}, \bibinfo{person}{Qin Han},
  \bibinfo{person}{Marc Habermann}, \bibinfo{person}{Kostas Daniilidis},
  \bibinfo{person}{Christian Theobalt}, {and} \bibinfo{person}{Lingjie Liu}.}
  \bibinfo{year}{2022}\natexlab{a}.
\newblock \bibinfo{title}{NeuS2: Fast Learning of Neural Implicit Surfaces for
  Multi-view Reconstruction}.
\newblock
\newblock


\bibitem[Wang et~al\mbox{.}(2022b)]%
        {wang2022hfneus}
\bibfield{author}{\bibinfo{person}{Yiqun Wang}, \bibinfo{person}{Ivan
  Skorokhodov}, {and} \bibinfo{person}{Peter Wonka}.}
  \bibinfo{year}{2022}\natexlab{b}.
\newblock \showarticletitle{HF-NeuS: Improved Surface Reconstruction Using
  High-Frequency Details}.
\newblock \bibinfo{journal}{\emph{arXiv preprint arXiv:2206.07850}}
  (\bibinfo{year}{2022}).
\newblock


\bibitem[Wang et~al\mbox{.}(2023)]%
        {wang2023fegr}
\bibfield{author}{\bibinfo{person}{Zian Wang}, \bibinfo{person}{Tianchang
  Shen}, \bibinfo{person}{Jun Gao}, \bibinfo{person}{Shengyu Huang},
  \bibinfo{person}{Jacob Munkberg}, \bibinfo{person}{Jon Hasselgren},
  \bibinfo{person}{Zan Gojcic}, \bibinfo{person}{Wenzheng Chen}, {and}
  \bibinfo{person}{Sanja Fidler}.} \bibinfo{year}{2023}\natexlab{}.
\newblock \showarticletitle{Neural Fields meet Explicit Geometric
  Representations for Inverse Rendering of Urban Scenes}. In
  \bibinfo{booktitle}{\emph{The IEEE Conference on Computer Vision and Pattern
  Recognition (CVPR)}}.
\newblock


\bibitem[Wood et~al\mbox{.}(2000)]%
        {wood2000surfacelf}
\bibfield{author}{\bibinfo{person}{Daniel~N. Wood}, \bibinfo{person}{Daniel~I.
  Azuma}, \bibinfo{person}{Ken Aldinger}, \bibinfo{person}{Brian Curless},
  \bibinfo{person}{Tom Duchamp}, \bibinfo{person}{David~H. Salesin}, {and}
  \bibinfo{person}{Werner Stuetzle}.} \bibinfo{year}{2000}\natexlab{}.
\newblock \showarticletitle{Surface Light Fields for 3D Photography}. In
  \bibinfo{booktitle}{\emph{Proceedings of the 27th Annual Conference on
  Computer Graphics and Interactive Techniques}}
  \emph{(\bibinfo{series}{SIGGRAPH '00})}. \bibinfo{publisher}{ACM
  Press/Addison-Wesley Publishing Co.}, \bibinfo{pages}{287–296}.
\newblock
\showISBNx{1581132085}


\bibitem[Xu and Harada(2022)]%
        {xu2022deforming}
\bibfield{author}{\bibinfo{person}{Tianhan Xu} {and} \bibinfo{person}{Tatsuya
  Harada}.} \bibinfo{year}{2022}\natexlab{}.
\newblock \showarticletitle{Deforming radiance fields with cages}. In
  \bibinfo{booktitle}{\emph{Computer Vision--ECCV 2022: 17th European
  Conference, Tel Aviv, Israel, October 23--27, 2022, Proceedings, Part
  XXXIII}}. Springer, \bibinfo{pages}{159--175}.
\newblock


\bibitem[Yariv et~al\mbox{.}(2021)]%
        {yariv2021volsdf}
\bibfield{author}{\bibinfo{person}{Lior Yariv}, \bibinfo{person}{Jiatao Gu},
  \bibinfo{person}{Yoni Kasten}, {and} \bibinfo{person}{Yaron Lipman}.}
  \bibinfo{year}{2021}\natexlab{}.
\newblock \showarticletitle{Volume rendering of neural implicit surfaces}. In
  \bibinfo{booktitle}{\emph{Thirty-Fifth Conference on Neural Information
  Processing Systems}}.
\newblock


\bibitem[Yariv et~al\mbox{.}(2023)]%
        {yariv2023bakedsdf}
\bibfield{author}{\bibinfo{person}{Lior Yariv}, \bibinfo{person}{Peter Hedman},
  \bibinfo{person}{Christian Reiser}, \bibinfo{person}{Dor Verbin},
  \bibinfo{person}{Pratul~P Srinivasan}, \bibinfo{person}{Richard Szeliski},
  \bibinfo{person}{Jonathan~T Barron}, {and} \bibinfo{person}{Ben Mildenhall}.}
  \bibinfo{year}{2023}\natexlab{}.
\newblock \showarticletitle{BakedSDF: Meshing Neural SDFs for Real-Time View
  Synthesis}.
\newblock \bibinfo{journal}{\emph{arXiv preprint arXiv:2302.14859}}
  (\bibinfo{year}{2023}).
\newblock


\bibitem[Yariv et~al\mbox{.}(2020)]%
        {yariv2020idr}
\bibfield{author}{\bibinfo{person}{Lior Yariv}, \bibinfo{person}{Yoni Kasten},
  \bibinfo{person}{Dror Moran}, \bibinfo{person}{Meirav Galun},
  \bibinfo{person}{Matan Atzmon}, \bibinfo{person}{Basri Ronen}, {and}
  \bibinfo{person}{Yaron Lipman}.} \bibinfo{year}{2020}\natexlab{}.
\newblock \showarticletitle{Multiview Neural Surface Reconstruction by
  Disentangling Geometry and Appearance}.
\newblock \bibinfo{journal}{\emph{Advances in Neural Information Processing
  Systems}}  \bibinfo{volume}{33} (\bibinfo{year}{2020}).
\newblock


\bibitem[Yu et~al\mbox{.}(2021)]%
        {yu2021plenoctrees}
\bibfield{author}{\bibinfo{person}{Alex Yu}, \bibinfo{person}{Ruilong Li},
  \bibinfo{person}{Matthew Tancik}, \bibinfo{person}{Hao Li},
  \bibinfo{person}{Ren Ng}, {and} \bibinfo{person}{Angjoo Kanazawa}.}
  \bibinfo{year}{2021}\natexlab{}.
\newblock \showarticletitle{{PlenOctrees} for Real-time Rendering of Neural
  Radiance Fields}. In \bibinfo{booktitle}{\emph{ICCV}}.
\newblock


\bibitem[Yuan et~al\mbox{.}(2022)]%
        {yuan2022nerf}
\bibfield{author}{\bibinfo{person}{Yu-Jie Yuan}, \bibinfo{person}{Yang-Tian
  Sun}, \bibinfo{person}{Yu-Kun Lai}, \bibinfo{person}{Yuewen Ma},
  \bibinfo{person}{Rongfei Jia}, {and} \bibinfo{person}{Lin Gao}.}
  \bibinfo{year}{2022}\natexlab{}.
\newblock \showarticletitle{NeRF-editing: geometry editing of neural radiance
  fields}. In \bibinfo{booktitle}{\emph{Proceedings of the IEEE/CVF Conference
  on Computer Vision and Pattern Recognition}}. \bibinfo{pages}{18353--18364}.
\newblock


\bibitem[Zhang et~al\mbox{.}(2021)]%
        {zhang2021PhySG}
\bibfield{author}{\bibinfo{person}{Kai Zhang}, \bibinfo{person}{Fujun Luan},
  \bibinfo{person}{Qianqian Wang}, \bibinfo{person}{Kavita Bala}, {and}
  \bibinfo{person}{Noah Snavely}.} \bibinfo{year}{2021}\natexlab{}.
\newblock \showarticletitle{{PhySG}: {I}nverse Rendering with Spherical
  Gaussians for Physics-based Material Editing and Relighting}. In
  \bibinfo{booktitle}{\emph{The IEEE/CVF Conference on Computer Vision and
  Pattern Recognition (CVPR)}}.
\newblock


\bibitem[Zhang et~al\mbox{.}(2020)]%
        {kaizhang2020nerf++}
\bibfield{author}{\bibinfo{person}{Kai Zhang}, \bibinfo{person}{Gernot
  Riegler}, \bibinfo{person}{Noah Snavely}, {and} \bibinfo{person}{Vladlen
  Koltun}.} \bibinfo{year}{2020}\natexlab{}.
\newblock \showarticletitle{NeRF++: Analyzing and Improving Neural Radiance
  Fields}.
\newblock \bibinfo{journal}{\emph{arXiv:2010.07492}} (\bibinfo{year}{2020}).
\newblock


\bibitem[Zhao et~al\mbox{.}(2022)]%
        {Zhao2022neuralAM}
\bibfield{author}{\bibinfo{person}{Fuqiang Zhao}, \bibinfo{person}{Yuheng
  Jiang}, \bibinfo{person}{Kaixin Yao}, \bibinfo{person}{Jiakai Zhang},
  \bibinfo{person}{Liao Wang}, \bibinfo{person}{Haizhao Dai},
  \bibinfo{person}{Yuhui Zhong}, \bibinfo{person}{Yingliang Zhang},
  \bibinfo{person}{Minye Wu}, \bibinfo{person}{Lan Xu}, {and}
  \bibinfo{person}{Jingyi Yu}.} \bibinfo{year}{2022}\natexlab{}.
\newblock \showarticletitle{Human Performance Modeling and Rendering via Neural
  Animated Mesh}.
\newblock \bibinfo{journal}{\emph{ACM Trans. Graph.}} \bibinfo{volume}{41},
  \bibinfo{number}{6}, Article \bibinfo{articleno}{235} (\bibinfo{year}{2022}),
  \bibinfo{numpages}{17}~pages.
\newblock


\end{thebibliography}
}

\appendix
\section*{Appendix}
We provide additional algorithmic details and pseudocode.
In the first phase of training, our method adaptively extracts an explicit mesh envelope which spatially bounds the neural volumetric representation: level set evolution and shell extraction are shown in Procedures~\ref{proc:levelset} and~\ref{proc:shellextraction}.
In the second phase of training, as well as inference, we leverage the extracted shells to sample query points only where they are needed. We cast rays against the shell meshes and compute query locations in the narrow band between the outer and inner shell. This is detailed in Procedure~\ref{proc:sampling}. Note that one ray may intersect with multiple narrow bands, however it always terminates when encountering an inner shell.
Finally, we include the overall training pipeline in Procedure~\ref{proc:training}.

\noindent
\begin{minipage}{\columnwidth} %
\vspace{2em}
\begin{algo}{\Proc{LevelSetEvolution}$(\SDF, \LevelsetVelocity_\text{evolve}, T, \Delta t, \zeta, \lambda_\textrm{curv})$}
  \label{proc:levelset}
  \begin{algorithmic}[1]
    \InputConditions{
        level set field $\SDF \in \mathbb{R}^{X \times Y \times Z}$, velocity $v_\text{evolve} \in \mathbb{R}^{X \times Y \times Z}$, evolution steps $T$, timestep $\Delta t$, soft falloff threshold $\zeta$, curvature regularization weight $\lambda_\textrm{curv}$
    }
    \OutputConditions{
        Evolved level set field $\SDF_T$
    }
    \State $\SDF_{0} \gets \SDF$ \Comment{Initialize level set field}
    \For{$i \gets 0$ to $T-1$}
        \State $\Big[\frac{\partial \SDF_i}{\partial t}\Big]_\textrm{mot} \gets -\left|\nabla \SDF_i \right| \LevelsetVelocity_\text{evolve}$ \Comment{Motion term}
        \State $\Big[\frac{\partial \SDF_i}{\partial t}\Big]_\textrm{curv} \gets -\lambda_\textrm{curv} \left|\nabla \SDF_i \right| \big(\nabla \cdot \dfrac{\nabla \SDF}{|\nabla \SDF|} \big)$ \Comment{Curvature term}
        \State $\omega_i \gets \tfrac{1}{2}\big(1 + \cos(\pi\ \textrm{clamp}(\SDF_i / \zeta, -1., 1.)\big)$ \Comment{Soft falloff (Eq.~\ref{eq:falloff})}
        \State $\frac{\partial \SDF_i}{\partial t} \gets \omega_i \,\big(\Big[\frac{\partial \SDF_i}{\partial t}\Big]_\textrm{mot} + \Big[\frac{\partial \SDF_i}{\partial t}\Big]_\textrm{curv} \big)$

        \State ${\SDF}_{i+1} = {\SDF}_i + \Delta t \,\frac{\partial \SDF_i}{\partial t}$  \Comment{Update level set field}
    \EndFor
    \State \textbf{Return} $\SDF_T$
  \end{algorithmic}
\end{algo}
\end{minipage}

\noindent
\begin{minipage}{\columnwidth} %
\vspace{1.5em}
\begin{algo}{\Proc{ShellExtraction}$(\SDF, \KernelSize, \tau_d, \DensityDilateVelCoef, \DensityMin, \DensityErodeVelCoef, \ErodeVelocityMax)$}
  \label{proc:shellextraction}
  \begin{algorithmic}[1]
    \InputConditions{
        signed distance field $\SDF \in \mathbb{R}^{X \times Y \times Z}$, spatially-varying kernel size $\KernelSize \in \mathbb{R}^{X \times Y \times Z}$,
        grid size $\tau_d$, dilation hyperparameters $\DensityDilateVelCoef, \DensityMin$, erosion hyperparameters $\DensityErodeVelCoef, \ErodeVelocityMax$
    }
    \OutputConditions{
        outer mesh $M_{+}$ and inner mesh $M_{-}$
    }
    \State {$\alpha \gets \frac{\textrm{Sigmoid}\big( (\SDF - \tau_d / 2) /\KernelSize \big) - \textrm{Sigmoid}\big( (\SDF + \tau_d / 2) /\KernelSize \big)}{\textrm{Sigmoid}\big( (\SDF - \tau_d / 2) /\KernelSize \big)}$} \LeftComment{Eq.~\ref{eq:neus_remapping}, as in NeuS}
    \\
    \State {}
    \LeftComment{Level set dilation for outer mesh}
    \State {$\LevelsetVelocity_\textrm{dilate} \gets \text{Where}(\alpha > \DensityMin, \DensityDilateVelCoef \alpha, 0)$ \Comment{Eq.~\ref{eq:dilation}}}
    \State {$\SDF_\textrm{dilate} \gets \text{LevelSetEvolution}(\SDF, \LevelsetVelocity_\textrm{dilate}, T=50, \Delta t=0.1, $}
    \State {$\hspace{11.6em} \zeta=0.1, \lambda_\textrm{curv}=0.01$)}
    \State {$\SDF_\textrm{dilate} \gets \min(\SDF_0, \SDF_\textrm{dilate})$ \Comment{Clip SDF for a strict dilation}}
    \State {$M_{+} \gets \text{MarchingCubes}(\SDF_\textrm{dilate})$}
    \\
    \State {}
    \LeftComment{Level set erosion for inner mesh}
    \State {$\LevelsetVelocity_\textrm{erode} \gets \min(\ErodeVelocityMax, \DensityErodeVelCoef / \alpha)$ \Comment{Eq.~\ref{eq:erosion}}}
    \State {$\SDF_\textrm{erode} \gets \text{LevelSetEvolution}(\SDF, \LevelsetVelocity_\textrm{erode}, T=50, \Delta t=0.1, $}
    \State {$\hspace{11.6em} \zeta=0.05, \lambda_\textrm{curv}=0$)}
    \State $\SDF_\textrm{erode} \gets \max(\SDF_0, \SDF_\textrm{erode})$ \Comment{Clip SDF for a strict erosion}
    \State {$M_{-} \gets \text{MarchingCubes}(\SDF_\textrm{erode})$}
    \\
    \State \textbf{Return} $M_{+}, M_{-}$
  \end{algorithmic}
\end{algo}
\vspace{1em}
\end{minipage}

\noindent
\begin{minipage}{\columnwidth} %
\vspace{2em}
\begin{algo}{\Proc{NarrowBandSampling}$(M_{+}, M_{-}, r, o, w_s, \delta_s,$ \\
                              $\color{white}{.} \hspace{5.05cm} \color{black}{N_{max}, dp_{max})}$}
  \label{proc:sampling}
  \begin{algorithmic}[1]
    \InputConditions{outer mesh $M_{+}$ and inner mesh $M_{-}$, ray origin $\mathbf{o} \in \mathbb{R}^3$ and direction $\mathbf{r} \in \mathbb{R}^3$, target inter-sample spacing $\delta_s$, single-sample threshold $w_s$, maximum number of samples per interval $N_{max}$, and a maximum cap for depth peeling $dp_{max}$}
    \OutputConditions{a list of distances $\tau_s$ to sampled points along the ray}

    \State $\tau_s \gets \text{empty list}$, \ $n_{hits} \gets 0$  \Comment{Initialize Steps}
    \\
    \LeftComment{Find hits of the inner mesh}
    \State $rayHits_{M_{-}} \gets $ CastRay($M_{-}, \mathbf{o}, \mathbf{r}$)
        \If{$\text{HasNext}(rayHits_{M_{-}})$}
           \State $(d_{M_{-}}, flag)\gets \text{GetNextHit}(rayHits_{M_{-}})$
        \Else
            \State $d_{M_{-}} \gets \infty$
        \EndIf
    \\
    \LeftComment{Loop through the hits of the outer mesh}
    \State $rayHits_{M_{+}} \gets $ CastRay($M_{+}, \mathbf{o}, \mathbf{r}$)
    \While{$\text{HasNext}(rayHits_{M_{+}})$ AND $n_{hits} < dp_{max}$}
        \State $(hitDistance, flag) \gets \text{GetNextHit}(rayHits_{M_{+}})$
        \State $n_{hits} \gets n_{hits}+1$
        \If{$flag = \text{ENTERING}$} \Comment{Ray enters the mesh band}
            \State $d_{enter} \gets$ $hitDistance$
        \ElsIf{$flag = \text{EXITING}$} \Comment{Ray exits the mesh band}
            \State {\LeftComment{Compute samples between enter and exit}}
            \State $d_{exit} \gets$ Min($hitDistance$, $d_{M_{-}}$)
            \State $w \gets d_{exit} - d_{enter}$
            \State $N \gets $ Min(Ceil(Max($w-w_s,0$)/$\delta_s$)+1,$N_{max}$)
            \State $\tau_s \gets \tau_s + \text{Linspace}(d_{enter}, d_{exit}, N+2)[1:-1]$
            \If{$hitDistance$  $> d_{M_{-}}$}
            \State \textbf{break} \Comment{Terminate if beyond the inner mesh}
            \EndIf
        \EndIf
    \EndWhile
    \State \textbf{Return} $\tau_s$
  \end{algorithmic}
\end{algo}
\end{minipage}

\noindent
\begin{minipage}{\columnwidth} %
\vspace{1.5em}
\begin{algo}{\Proc{Training Pipeline}}
  \label{proc:training}
  \begin{algorithmic}[1]
    \InputConditions{rays $\mathcal{R}$, ground-truth pixel colors $\mathcal{C}$, training iterations for the first stage $N_1$ and second stage $N_2$, network $\GenericNetwork$}
    \OutputConditions{optimized network parameters $\GenericNetworkParameters$, shell $M_{+}, M_{-}$}
    \LeftComment{First stage training with full ray volume rendering}
    \For{$i \gets 0$ to $N_1-1$}
        \State Sample data $\textbf{r}_i \in \mathcal{R}, \textbf{c}_i \in \mathcal{C}$
        \State $output \gets \text{VolumeRendering}(\GenericNetwork, \textbf{r}_i)$
        \State $\Loss \gets \text{Loss}(output, \textbf{c}_i)$ \Comment{Compute loss with Eq.~\ref{eq:loss}}
        \State Update network: $\GenericNetworkParameters \gets \GenericNetworkParameters - \eta \frac{\partial \Loss}{\partial \GenericNetworkParameters}$
    \EndFor
    \\
    \LeftComment{Extract adaptive shells}
    \State $M_{+}, M_{-} \gets \text{ShellExtraction}(\GenericNetwork)$ \Comment{\secref{shell_extraction}}
    \\
    \LeftComment{Second stage training with narrow-band rendering}
    \For{$i \gets 0$ to $N_2-1$}
        \State Sample data $\textbf{r}_i \in \mathcal{R}, \textbf{c}_i \in \mathcal{C}$
        \State $output \gets \text{NarrowBandRendering}(\GenericNetwork, \textbf{r}_i, M_{+}, M_{-})$
        \State $\Loss_\Color \gets \text{ColorLoss}(output, \textbf{c}_i)$ \Comment{Compute loss with Eq.~\ref{eq:loss_color}}
        \State Update network: $\GenericNetworkParameters \gets \GenericNetworkParameters - \eta \frac{\partial \Loss_\Color}{\partial \GenericNetworkParameters}$
    \EndFor
    \State \textbf{Return} $\GenericNetworkParameters$, $M_{+}, M_{-}$
  \end{algorithmic}
\end{algo}
\end{minipage}

\end{document}


\title{Adaptive Shells for Efficient Neural Radiance Field Rendering
\\
Supplementary Material}

\acmSubmissionID{774}

\keywords{}

\begin{abstract}
\end{abstract}
\maketitle

In this supplementary material, we provide additional results~(\secref{additional_results}) and implementation details (\secref{implementation_details}).

\section{Additional results}
\label{sec:additional_results}
\begin{figure*}
    \centering
    \includegraphics[width=0.98\textwidth]{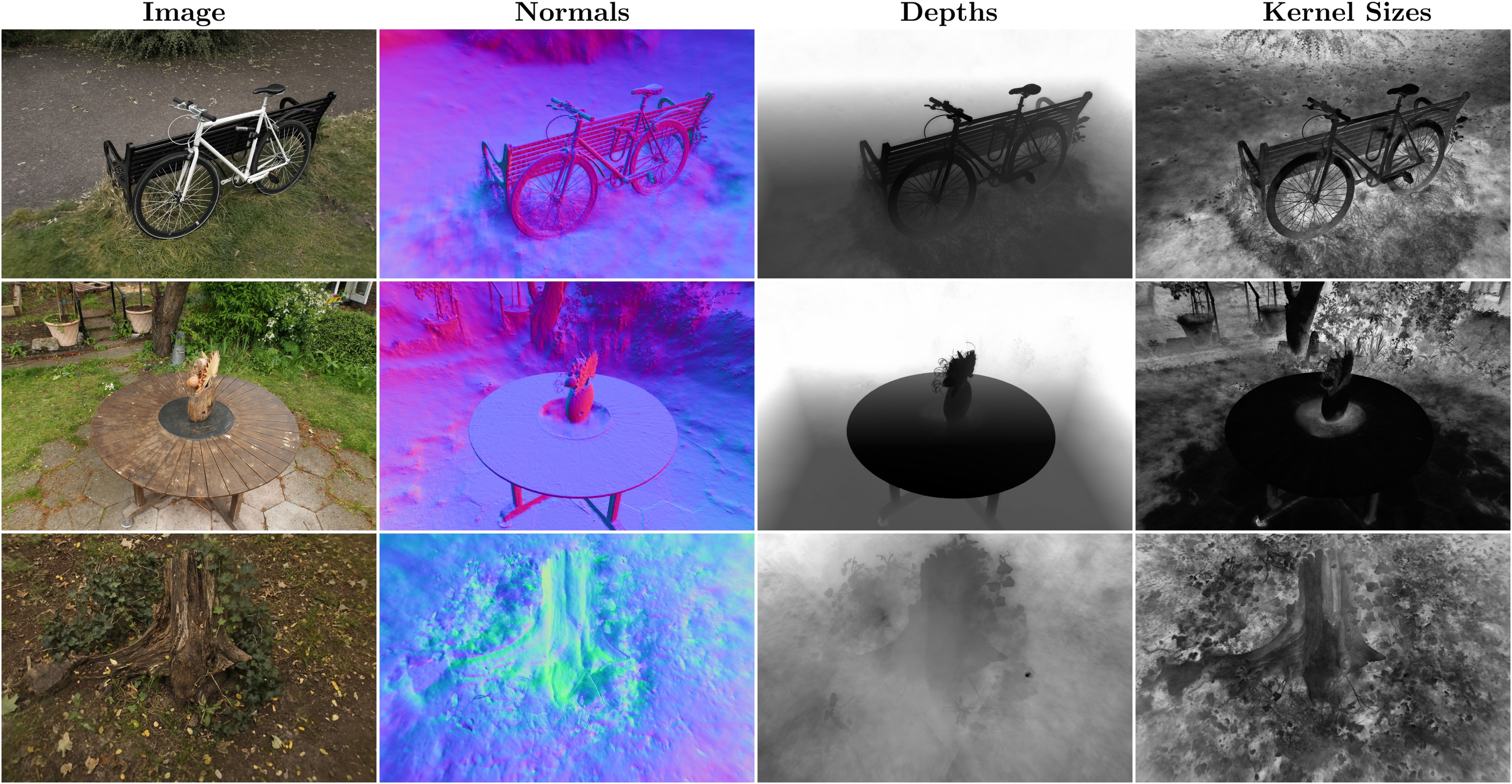}
    \vspace*{-1em}
    \caption{
        Qualitative visualization of geometry and kernel size on \MipNerfDataset{} data set. The kernel size is re-scaled to between 0 and 1. Our method automatically converges to a large kernel size for fuzzy regions such as grass and a small kernel size for sharp surfaces.
        \label{fig:gallery_mipnerf360_stage1}
    }
\end{figure*}

We provide additional qualitative results on \MipNerfDataset{} data set (\figref{gallery_mipnerf360_stage1}) and \DTUDataset{} data set (Figure~\ref{fig:gallery_dtu_full_1}, \ref{fig:gallery_dtu_full_2}), as well as per-scene quantitative numbers for all data sets (Tables~\ref{tab:shelly_details_psnr}--\ref{tab:mipnerf_details_sample}).

\section{Implementation details}
\label{sec:implementation_details}
\paragraph{Level set evolution}
To perform the level set evolution, we extract the initial SDF values on a $512^3$ grid and separately dilate and erode the zero level set for 50 iterations with the timestep as $0.1$.
For dilation, we use $\DensityDilateVelCoef=1$ and the density threshold $\sigma_{\min}$ is set to $0.01$ for all data sets. These values were determined empirically such that the dilated level set sufficiently covers thin structures. We set the erosion hyperparameters to $\DensityErodeVelCoef=0.001, \ErodeVelocityMax=100$. The evolution process takes approximately 2 seconds which is negligible compared to the other steps of our training pipeline.

\paragraph{Narrow-band rendering}
When sampling query points within the shells, we set the maximum number of samples per interval as $N_{max}=16$, and the maximum cap for depth peeling $dp_{max}=20$.
For \ShellyDataset{}, \DTUDataset{} and \MipNerfDataset{} data sets, we use single-sample threshold $w_s=0.02$ and inter-sample spacing $\delta_s=0.01$. For \NeRFDataset{} data set, we use single-sample threshold $w_s=0.005$ and inter-sample spacing $\delta_s=0.0025$.

\paragraph{Representing the background}
For \DTUDataset{} data set, we combine the main volume representation with a spherical background placed at infinity (only dependent on the ray direction). Similar to our main network, the spherical background is represented using a combination of a hash encoding (4 levels, 2D features per level) and a small MLP such that $\Color = \BackgroundNetwork([\PosEncoding(\Direction)])$. All the rays that completely miss the extracted shell obtain the color from the background, which requires a single sample evaluation.

When training on \MipNerfDataset{} data set, we follow prior works \cite{yariv2023bakedsdf} and extend our method with the scene contraction proposed in \cite{barron2022mipnerf360}. Specifically, we map the scene outside the unit sphere into a sphere with radius 2 using the scene contraction function
\begin{equation}
    \text{contract}(\textbf{x}) = \left\{
    \begin{aligned}
        & \textbf{x} , &||\textbf{x}|| \leq 1, \\
        & (2 - \frac{1}{||\textbf{x}||}) \frac{\textbf{x}}{||\textbf{x}||},  &||\textbf{x}|| > 1.
    \end{aligned}
    \right.
\end{equation}

\paragraph{Training details.}
During training, we linearly warm up the learning rate to $1\times 10^{-2}$ in the first 5k iterations, and then exponentially decay it to $1\times 10^{-4}$ at the end of training. In all experiments, we use the AdamW optimizer with weight decay as 1e-2. For \ShellyDataset{}, \DTUDataset{} and \NeRFDataset{} data sets, we train for a total of 300k iterations, where the first 200k iterations are used for the first stage (full-ray formulation) and the remaining 100k for the second stage (finetuning within the narrow band). We use the batch-size of 4096 rays. For each scene, the training takes 8h on a single A100 GPU. For larger \MipNerfDataset{} scenes, we increase the full-ray training to 500k iterations.  We adopt the progressive training scheme~\cite{wang2022neus2,Li2023Neuralangelo}, where we initially enable the 8 coarsest levels of features (with the remaining feature channels set to zero), we then add one level every 5k iterations until reaching the maximum number of levels which equals 14.

\begin{table*}[!thbp]
\centering
\small
\caption{\revAdded{Per-scene quantitative PSNR comparison on \ShellyDataset{} data set. }}
\label{tab:shelly_details_psnr}
\vspace{-1em}
\begin{tabular}{c|c|c|c|c|c|c|c}
\toprule
PSNR~$\uparrow$ & \tabincell{c}{NeRF\\~\cite{mildenhall2020nerf}} & \tabincell{c}{MipNeRF\\\cite{barron2021mipnerf}} & \tabincell{c}{NeuS\\\cite{wang2021neus}} & \tabincell{c}{I-NGP\\\cite{mueller2022ingp}} & \tabincell{c}{MobileNeRF\\\cite{chen2022mobilenerf}} & \tabincell{c}{Ours\\(full ray)} & {Ours} \\
\midrule
Fernvase & 31.77 & 32.54 & 29.26 & 33.48 & 31.38 & 33.93 & 36.47 \\
Pug      & 31.34 & 32.26 & 31.38 & 32.70 & 31.50 & 33.60 & 35.83 \\
Woolly   & 28.33 & 29.18 & 27.90 & 31.13 & 31.61 & 31.61 & 34.19 \\
Horse 	 & 34.02 & 37.12 & 30.92 & 37.67 & 36.48 & 39.40 & 40.57 \\
Khady 	 & 29.01 & 29.88 & 28.29 & 29.21 & 26.84 & 31.09 & 31.22 \\
Kitten   & 33.18 & 34.54 & 32.16 & 35.13 & 31.93 & 35.94 & 37.82 \\
\midrule
Average  & 31.28 & 32.59 & 29.98 & 33.22 & 31.62 & 34.26 & 36.02 \\
\bottomrule
\end{tabular}
\end{table*}

\begin{table*}[!thbp]
\centering
\small
\caption{\revAdded{Per-scene quantitative LPIPS comparison on \ShellyDataset{} data set. }}
\label{tab:shelly_details_lpips}
\vspace{-1em}
\begin{tabular}{c|c|c|c|c|c|c|c}
\toprule
LPIPS~$\downarrow$ & \tabincell{c}{NeRF\\~\cite{mildenhall2020nerf}} & \tabincell{c}{MipNeRF\\\cite{barron2021mipnerf}} & \tabincell{c}{NeuS\\\cite{wang2021neus}} & \tabincell{c}{I-NGP\\\cite{mueller2022ingp}} & \tabincell{c}{MobileNeRF\\\cite{chen2022mobilenerf}} & \tabincell{c}{Ours\\(full ray)} & {Ours} \\
\midrule
Fernvase & 0.093 & 0.088 & 0.094 & 0.068 & 0.074 & 0.065 & 0.046 \\
Pug      & 0.198 & 0.190 & 0.209 & 0.156 & 0.167 & 0.132 & 0.093 \\
Woolly   & 0.241 & 0.217 & 0.232 & 0.181 & 0.163 & 0.139 & 0.089 \\
Horse 	 & 0.071 & 0.064 & 0.067 & 0.049 & 0.057 & 0.036 & 0.029 \\
Khady 	 & 0.246 & 0.239 & 0.251 & 0.215 & 0.218 & 0.185 & 0.160 \\
Kitten   & 0.094 & 0.092 & 0.097 & 0.079 & 0.094 & 0.066 & 0.056 \\
\midrule
Average  & 0.157 & 0.148 & 0.158 & 0.125 & 0.129 & 0.104 & 0.079 \\
\bottomrule
\end{tabular}
\end{table*}

\begin{table*}[!thbp]
\centering
\small
\caption{\revAdded{Per-scene quantitative SSIM comparison on \ShellyDataset{} dataset. }}
\label{tab:shelly_details_ssim}
\vspace{-1em}
\begin{tabular}{c|c|c|c|c|c|c|c}
\toprule
SSIM~$\uparrow$ & \tabincell{c}{NeRF\\~\cite{mildenhall2020nerf}} & \tabincell{c}{MipNeRF\\\cite{barron2021mipnerf}} & \tabincell{c}{NeuS\\\cite{wang2021neus}} & \tabincell{c}{I-NGP\\\cite{mueller2022ingp}} & \tabincell{c}{MobileNeRF\\\cite{chen2022mobilenerf}} & \tabincell{c}{Ours\\(full ray)} & {Ours} \\
\midrule
Fernvase & 0.937 & 0.940 & 0.932 & 0.955 & 0.944 & 0.964 & 0.976 \\
Pug      & 0.863 & 0.868 & 0.865 & 0.896 & 0.885 & 0.910 & 0.947 \\
Woolly   & 0.805 & 0.822 & 0.803 & 0.876 & 0.891 & 0.896 & 0.950 \\
Horse 	 & 0.975 & 0.980 & 0.973 & 0.985 & 0.980 & 0.988 & 0.992 \\
Khady 	 & 0.831 & 0.835 & 0.833 & 0.852 & 0.823 & 0.862 & 0.881 \\
Kitten   & 0.949 & 0.954 & 0.949 & 0.967 & 0.942 & 0.969 & 0.979 \\
\midrule
Average  & 0.893 & 0.899 & 0.893 & 0.922 & 0.911 & 0.932 & 0.954 \\
\bottomrule
\end{tabular}
\end{table*}

\begin{table*}[!thbp]
\centering
\small
\caption{\revAdded{Per-scene sample count of our method on \ShellyDataset{} data set. }}
\label{tab:shelly_details_sample}
\vspace{-1em}
\begin{tabular}{c|c|c|c|c|c|c|c}
\toprule
 & Fernvase & Pug & Woolly & Horse & Khady & Kitten & Average \\
\midrule
Ours  & 1.514 & 1.921 & 2.040 & 0.425 & 3.292 & 1.228 & 1.737 \\
\bottomrule
\end{tabular}
\end{table*} 

\begin{table*}[!thbp]
\centering
\small
\caption{Per-scene quantitative results on \DTUDataset{} data set. We report the PSNR, LPIPS and SSIM results for each scene and compare them with baselines. \revRemoved{Our method achieves much better performance compared with baselines.}}
\label{tab:dtu_full_results}
\vspace{-1em}
\begin{adjustbox}{width=\textwidth,center}
\begin{tabular}{c|ccc|ccc|ccc|ccc|ccc|ccc}
\toprule
& \multicolumn{3}{c|}{I-NGP~\cite{mueller2022ingp}} & \multicolumn{3}{c|}{Neus~\cite{wang2021neus}} & \multicolumn{3}{c|}{Nerf~\cite{mildenhall2020nerf}} & \multicolumn{3}{c|}{MipNeRF~\cite{barron2021mipnerf}} & \multicolumn{3}{c|}{Ours (full ray)} & \multicolumn{3}{c}{Ours} \\
\midrule
Scene & PSNR~$\uparrow$ & LPIPS~$\downarrow$ & SSIM~$\uparrow$ &PSNR~$\uparrow$ & LPIPS~$\downarrow$ & SSIM~$\uparrow$ &PSNR~$\uparrow$ & LPIPS~$\downarrow$ & SSIM~$\uparrow$ &PSNR~$\uparrow$ & LPIPS~$\downarrow$ & SSIM~$\uparrow$ &PSNR~$\uparrow$ & LPIPS~$\downarrow$ & SSIM~$\uparrow$ &PSNR~$\uparrow$ & LPIPS~$\downarrow$ & SSIM~$\uparrow$  \\
\midrule
24 & 29.85 & 0.237 & 0.871 & 26.22 & 0.324 & 0.787 & 24.87 & 0.364 & 0.751 & 25.60 & 0.359 & 0.763 & 32.42 & 0.110 & 0.931 & 30.91 & 0.102 & 0.934 \\
37 & 25.05 & 0.169 & 0.869 & 23.63 & 0.191 & 0.835 & 21.31 & 0.225 & 0.792 & 22.97 & 0.212 & 0.815 & 27.46 & 0.095 & 0.932 & 27.38 & 0.096 & 0.938 \\
40 & 28.80 & 0.291 & 0.829 & 26.38 & 0.344 & 0.755 & 24.98 & 0.363 & 0.715 & 25.90 & 0.343 & 0.743 & 31.16 & 0.161 & 0.913 & 32.10 & 0.150 & 0.931 \\
55 & 27.80 & 0.171 & 0.923 & 25.56 & 0.171 & 0.893 & 23.61 & 0.202 & 0.846 & 24.20 & 0.192 & 0.865 & 32.09 & 0.067 & 0.974 & 32.35 & 0.065 & 0.975 \\
63 & 33.06 & 0.075 & 0.961 & 30.51 & 0.115 & 0.954 & 29.94 & 0.117 & 0.945 & 30.41 & 0.114 & 0.947 & 34.94 & 0.049 & 0.974 & 34.10 & 0.044 & 0.975 \\
65 & 33.75 & 0.091 & 0.962 & 30.71 & 0.108 & 0.962 & 29.86 & 0.118 & 0.948 & 30.38 & 0.114 & 0.951 & 34.90 & 0.069 & 0.069 & 35.08 & 0.066 & 0.974 \\
69 & 31.21 & 0.149 & 0.946 & 27.97 & 0.184 & 0.938 & 28.15 & 0.217 & 0.913 & 28.75 & 0.213 & 0.916 & 32.15 & 0.104 & 0.961 & 31.79 & 0.102 & 0.960 \\
83 & 35.28 & 0.055 & 0.977 & 33.57 & 0.081 & 0.974 & 33.13 & 0.077 & 0.971 & 33.29 & 0.077 & 0.971 & 36.75 & 0.040 & 0.983 & 37.12 & 0.036 & 0.985 \\
97 & 28.50 & 0.140 & 0.937 & 26.93 & 0.144 & 0.936 & 26.58 & 0.168 & 0.918 & 26.70 & 0.172 & 0.918 & 29.99 & 0.086 & 0.959 & 29.77 & 0.093 & 0.956 \\
105 & 34.08 & 0.113 & 0.948 & 31.39 & 0.164 & 0.932 & 31.37 & 0.160 & 0.927 & 31.35 & 0.160 & 0.926 & 35.50 & 0.067 & 0.967 & 35.91 & 0.061 & 0.970 \\
106 & 33.31 & 0.135 & 0.947 & 28.80 & 0.161 & 0.934 & 30.63 & 0.177 & 0.921 & 30.92 & 0.174 & 0.923 & 36.29 & 0.073 & 0.973 & 35.81 & 0.072 & 0.972 \\
110 & 29.89 & 0.118 & 0.951 & 28.14 & 0.145 & 0.946 & 28.84 & 0.140 & 0.943 & 28.55 & 0.140 & 0.942 & 32.81 & 0.073 & 0.971 & 33.18 & 0.071 & 0.972 \\
114 & 29.41 & 0.163 & 0.925 & 28.13 & 0.178 & 0.921 & 28.28 & 0.183 & 0.907 & 28.33 & 0.183 & 0.908 & 31.08 & 0.102 & 0.955 & 31.07 & 0.094 & 0.956 \\
118 & 35.23 & 0.105 & 0.964 & 31.60 & 0.128 & 0.961 & 33.42 & 0.130 & 0.952 & 33.23 & 0.129 & 0.951 & 37.67 & 0.066 & 0.978 & 36.71 & 0.062 & 0.979 \\
122 & 35.42 & 0.074 & 0.972 & 34.36 & 0.090 & 0.971 & 32.66 & 0.102 & 0.959 & 32.97 & 0.098 & 0.961 & 37.50 & 0.047 & 0.983 & 37.21 & 0.044 & 0.983 \\
\midrule
Mean & 31.38 & 0.139 & 0.932 & 28.93 & 0.168 & 0.913 & 28.51 & 0.183 & 0.894 & 28.90 & 0.179 & 0.900 & 33.51 & 0.081 & 0.901 & 33.37 & 0.077 & 0.964 \\
\bottomrule
\end{tabular}
\end{adjustbox}
\end{table*}

\begin{table*}[!thbp]
\centering
\small
\caption{\revAdded{Per-scene sample count of our method on \DTUDataset{} data set. }}
\label{tab:dtu_details_sample}
\vspace{-1em}
\revAdded{
\begin{tabular}{c|c|c|c|c|c|c|c|c|c|c|c|c|c|c|c|c}
\toprule
Scene & 24 & 37 & 40 & 55 & 63 & 65 & 69 & 83 & 97 & 105 & 106 & 110 & 114 & 118 & 122 & Average \\
\midrule
Ours  & 3.989 & 5.716 & 4.085 & 5.254 & 3.570 & 4.385 & 6.844 & 4.608 & 5.882 & 5.750 & 4.015 & 6.492 & 4.454 & 3.117 & 2.910 & 4.738 \\
\bottomrule
\end{tabular}
}
\end{table*}

\begin{table*}[!thbp]
\centering
\small
\caption{\revAdded{Per-scene results of our method on \NeRFDataset{} data set. }}
\label{tab:nerfsynthetic_details_sample}
\vspace{-1em}
\begin{tabular}{c|c|c|c|c|c|c|c|c|c}
\toprule
Ours (full ray)  & Mic & Ficus & Chair & Hotdog & Materials & Drums & Ship & Lego & Average \\
\midrule
PSNR  & 34.46 & 34.67 & 35.14 & 36.17 & 28.47 & 25.32 & 30.27 & 35.60 & 32.51 \\
LPIPS & 0.012 & 0.024 & 0.020 & 0.029 & 0.076 & 0.081 & 0.121 & 0.021 & 0.048 \\
SSIM  & 0.989 & 0.985 & 0.986 & 0.982 & 0.941 & 0.939 & 0.896 & 0.981 & 0.962 \\
\midrule
\midrule
Ours & Mic & Ficus & Chair & Hotdog & Materials & Drums & Ship & Lego & Average \\
\midrule
PSNR         & 33.91 & 33.63 & 34.94 & 36.21 & 27.82 & 25.19 & 29.54 & 33.49 & 31.84 \\
LPIPS        & 0.015 & 0.033 & 0.023 & 0.035 & 0.086 & 0.086 & 0.141 & 0.031 & 0.056 \\
SSIM         & 0.988 & 0.981 & 0.985 & 0.981 & 0.935 & 0.937 & 0.877 & 0.973 & 0.957 \\
Sample count & 1.200 & 3.097 & 2.113 & 3.728 & 4.235 & 3.042 & 6.799 & 4.035 & 3.531 \\
\bottomrule
\end{tabular}
\end{table*}

\begin{table*}[!thbp]
\centering
\small
\caption{\revAdded{Per-scene quantitative PSNR comparison on \MipNerfDataset{} data set. }}
\label{tab:mipnerf_details_psnr}
\vspace{-1em}
\begin{adjustbox}{width=\textwidth,center}
\revAdded{
\begin{tabular}{cc|c|c|c|c|c|c|c|c}
\toprule
& & \tabincell{c}{NeRF\\\cite{mildenhall2020nerf}} & \tabincell{c}{Mip-NeRF\\\cite{barron2021mipnerf}} & \tabincell{c}{Mip-NeRF 360\\\cite{barron2022mipnerf360}} & \tabincell{c}{I-NGP\\\cite{mueller2022ingp}} & \tabincell{c}{MobileNeRF\\\cite{chen2022mobilenerf}} & \tabincell{c}{BakedSDF\\\cite{yariv2023bakedsdf}} & \tabincell{c}{Ours \\(full ray)} & \tabincell{c}{Ours} \\
\midrule
\multirow{4}{*}{ Outdoor } 
& Bicycle & 21.76 & 21.69 & 24.37 & 23.67 & 21.70 & 22.08 & 24.07 & 22.19 \\
& Garden  & 23.11 & 23.16 & 26.98 & 24.60 & 23.04 & 24.53 & 25.73 & 25.35 \\
& Stump   & 21.73 & 21.21 & 26.40 & 23.43 & 23.96 & 23.59 & 23.10 & 21.96 \\
\midrule
& Average & 22.20 & 22.02 & 25.92 & 23.90 & 22.90 & 23.40 & 24.30 & 23.17 \\
\midrule
\midrule
\multirow{5}{*}{ Indoor } 
& Room    & 28.56 & 28.73 & 31.63 & 30.16 & 28.76 & 28.63 & 29.61 & 30.63 \\
& Counter & 25.67 & 25.59 & 29.55 & 26.03 & 24.74 & 25.63 & 26.26 & 25.24 \\
& Kitchen & 26.31 & 26.47 & 32.23 & 29.86 & 26.28 & 26.88 & 30.10 & 28.43 \\
& Bonsai  & 26.81 & 27.13 & 33.46 & 31.82 & 23.20 & 27.67 & 30.19 & 32.47 \\
\midrule
& Average & 26.84 & 26.98 & 31.72 & 29.47 & 25.74 & 27.20 & 29.04 & 29.19 \\
\bottomrule
\end{tabular}
}
\end{adjustbox}
\end{table*}

\begin{table*}[!thbp]
\centering
\small
\caption{\revAdded{Per-scene quantitative LPIPS comparison on \MipNerfDataset{} data set. }}
\label{tab:mipnerf_details_lpips}
\vspace{-1em}
\begin{adjustbox}{width=\textwidth,center}
\revAdded{
\begin{tabular}{cc|c|c|c|c|c|c|c|c}
\toprule
& & \tabincell{c}{NeRF\\\cite{mildenhall2020nerf}} & \tabincell{c}{Mip-NeRF\\\cite{barron2021mipnerf}} & \tabincell{c}{Mip-NeRF 360\\\cite{barron2022mipnerf360}} & \tabincell{c}{I-NGP\\\cite{mueller2022ingp}} & \tabincell{c}{MobileNeRF\\\cite{chen2022mobilenerf}} & \tabincell{c}{BakedSDF\\\cite{yariv2023bakedsdf}} & \tabincell{c}{Ours \\(full ray)} & \tabincell{c}{Ours} \\
\midrule
\multirow{4}{*}{ Outdoor } 
& Bicycle & 0.536 & 0.541 & 0.301 & 0.417 & 0.513 & 0.394 & 0.326 & 0.438 \\
& Garden  & 0.415 & 0.422 & 0.170 & 0.248 & 0.396 & 0.243 & 0.200 & 0.247 \\
& Stump   & 0.551 & 0.490 & 0.261 & 0.441 & 0.480 & 0.415 & 0.421 & 0.482 \\
\midrule
& Average & 0.501 & 0.484 & 0.244 & 0.369 & 0.463 & 0.351 & 0.316 & 0.389 \\
\midrule
\midrule
\multirow{5}{*}{ Indoor } 
& Room    & 0.353 & 0.346 & 0.211 & 0.300 & 0.423 & 0.310 & 0.295 & 0.300 \\
& Counter & 0.394 & 0.390 & 0.201 & 0.341 & 0.476 & 0.340 & 0.276 & 0.395 \\
& Kitchen & 0.335 & 0.336 & 0.127 & 0.224 & 0.393 & 0.267 & 0.168 & 0.219 \\
& Bonsai  & 0.398 & 0.370 & 0.176 & 0.227 & 0.522 & 0.281 & 0.217 & 0.225 \\
\midrule
& Average & 0.370 & 0.361 & 0.179 & 0.273 & 0.453 & 0.300 & 0.239 & 0.285 \\
\bottomrule
\end{tabular}
}
\end{adjustbox}
\end{table*}

\begin{table*}[!thbp]
\centering
\small
\caption{\revAdded{Per-scene quantitative SSIM comparison on \MipNerfDataset{} data set. }}
\label{tab:mipnerf_details_ssim}
\vspace{-1em}
\begin{adjustbox}{width=\textwidth,center}
\revAdded{
\begin{tabular}{cc|c|c|c|c|c|c|c|c}
\toprule
& & \tabincell{c}{NeRF\\\cite{mildenhall2020nerf}} & \tabincell{c}{Mip-NeRF\\\cite{barron2021mipnerf}} & \tabincell{c}{Mip-NeRF 360\\\cite{barron2022mipnerf360}} & \tabincell{c}{I-NGP\\\cite{mueller2022ingp}} & \tabincell{c}{MobileNeRF\\\cite{chen2022mobilenerf}} & \tabincell{c}{BakedSDF\\\cite{yariv2023bakedsdf}} & \tabincell{c}{Ours \\(full ray)} & \tabincell{c}{Ours} \\
\midrule
\multirow{4}{*}{ Outdoor } 
& Bicycle & 0.455 & 0.454 & 0.685 & 0.624 & 0.426 & 0.394 & 0.689 & 0.544 \\
& Garden  & 0.546 & 0.543 & 0.813 & 0.708 & 0.589 & 0.740 & 0.814 & 0.757 \\
& Stump   & 0.453 & 0.517 & 0.744 & 0.613 & 0.557 & 0.597 & 0.607 & 0.516 \\
\midrule
& Average & 0.485 & 0.505 & 0.747 & 0.648 & 0.524 & 0.577 & 0.703 & 0.606 \\
\midrule
\midrule
\multirow{5}{*}{ Indoor } 
& Room    & 0.843 & 0.851 & 0.913 & 0.886 & 0.836 & 0.872 & 0.892 & 0.895 \\
& Counter & 0.775 & 0.779 & 0.894 & 0.826 & 0.724 & 0.809 & 0.872 & 0.795 \\
& Kitchen & 0.749 & 0.745 & 0.920 & 0.869 & 0.751 & 0.825 & 0.904 & 0.866 \\
& Bonsai  & 0.792 & 0.818 & 0.941 & 0.925 & 0.716 & 0.875 & 0.933 & 0.933 \\
\midrule
& Average & 0.790 & 0.798 & 0.917 & 0.877 & 0.757 & 0.845 & 0.900 & 0.872 \\
\bottomrule
\end{tabular}
}
\end{adjustbox}
\end{table*}

\begin{table*}[!thbp]
\centering
\small
\caption{\revAdded{Per-scene sample count of our method on \MipNerfDataset{} data set. }}
\label{tab:mipnerf_details_sample}
\vspace{-1em}
\begin{tabular}{c|c|c|c|c|c|c|c|c}
\toprule
 & Bicycle & Garden & Stump & Room & Counter & Kitchen & Bonsai & Average \\
\midrule
Ours  & 16.33 & 14.05 & 22.81 & 17.14 & 22.24 & 14.49 & 14.80 & 17.41 \\
\bottomrule
\end{tabular}
\end{table*}

{\small
\bibliographystyle{ACM-Reference-Format}
\bibliography{egbib}
}